\documentclass[twocolumn]{IEEEtran}

\usepackage{graphicx}
\usepackage{epsfig}
\usepackage{amsmath}
\usepackage{bbm}
\usepackage{amssymb}
\usepackage{setspace}
\usepackage{wrapfig}
\usepackage{times,color}
\usepackage{caption}
\usepackage{subcaption}
\usepackage{algorithm}
\usepackage{algorithmic}
\usepackage{epstopdf}
\graphicspath{{./eps/}}
\usepackage{cite,footnote,xspace,syntonly,bm}
\usepackage{amsfonts}

\input{mysymbol.sty}

\def \A {{\mathbf{A}}}
\def \B {{\mathbf{B}}}

\def \I {{\mathbf{I}}}
\def \Y {{\mathbf{Y}}}
\def \X {{\mathbf{X}}}

\def \E {{\mathbf{E}}}

\def \M {{\mathbf{M}}}

\def \G {{\mathbf{G}}}

\def \V {{\mathbf{V}}}

\def \Phib {{\boldsymbol{\Phi}}}
\def \y {{\mathbf{y}}}
\def \x {{\mathbf{x}}}

\def \a {{\mathbf{a}}}
\def \b {{\mathbf{b}}}

\long\def\symbolfootnote[#1]#2{\begingroup
\def\thefootnote{\fnsymbol{footnote}}
\footnote[#1]{#2}\endgroup} \psfull

\begin{document}


\title{\huge Tracking Switched Dynamic Network Topologies \\ from Information Cascades$^\dag$}

\author{{\it Brian Baingana, \textit{Student Member}, \textit{IEEE} and Georgios~B.~Giannakis, \textit{Fellow}, \textit{IEEE}$^\ast$}}

\markboth{IEEE TRANSACTIONS OF SIGNAL PROCESSING (SUBMITTED)}
\maketitle \maketitle \symbolfootnote[0]{$\dag$ Work in this paper was supported by NSF grants 1343248, 1442686, 1514056, NIH Grant No.1R01GM104975-01, and AFOSR-MURI Grant No. FA9550-10-1-0567. This research has been co-financed by the European Union (European Social Fund - ESF), and Greek national funds through the Operational Program ``Education and Lifelong Learning'' of the National Strategic Reference
Framework (NSRF) - Research Funding Program: THALES. Investing in knowledge society through the ESF. Parts of the paper will appear in the \emph{Proc. of the IEEE Global Conference on Signal and Information Processing}, Orlando, Florida, December 14-16, 2015}
\symbolfootnote[0]{$\ast$ The authors are with the Dept.
of ECE and the Digital Technology Center, University of
Minnesota, 200 Union Street SE, Minneapolis, MN 55455. Tel/fax:
(612)626-7781/625-4583; Emails:
\texttt{\{baing011,georgios\}@umn.edu}}


\thispagestyle{empty}\addtocounter{page}{-1}
\begin{abstract}
Contagions such as the spread of popular news stories, or infectious diseases, propagate in cascades over dynamic networks with unobservable topologies. However, ``social signals'' such as product purchase time, or blog entry timestamps are measurable, and implicitly depend on the underlying topology, making it possible to track it over time. Interestingly, network topologies often ``jump'' between discrete states that may account for sudden changes in the observed signals. The present paper advocates a switched dynamic structural equation model to capture the topology-dependent cascade evolution, as well as the discrete states driving the underlying topologies. Conditions under which the proposed switched model is identifiable are established. Leveraging the edge sparsity inherent to social networks, a recursive $\ell_1$-norm regularized least-squares estimator is put forth to jointly track the states and network topologies. An efficient first-order proximal-gradient algorithm is developed to solve the resulting optimization problem. Numerical experiments on both synthetic data and real cascades measured over the span of one year are conducted, and test results corroborate the efficacy of the advocated approach.
\end{abstract}

\IEEEpeerreviewmaketitle

\section{Introduction}
\label{sec:introduction}
Information often spreads in cascades by following implicit links between nodes in real-world networks whose topologies may be unknown. For example the spread of viral news among blogs, or Internet memes over microblogging tools like \emph{Twitter}, are facilitated by the inherent connectivity of the world-wide web.  A celebrity may post an interesting \emph{tweet}, which is then read and \emph{retweeted} by her followers. An information cascade may then emerge if the followers of her followers share the tweet, and so on. The dynamics of propagation of such information over implicit networks are remarkably similar to those that govern the rapid spread of infectious diseases, leading to the so-termed \emph{contagions}~\cite{rogers_book,easley_book,baingana_jstsp}. Similar cascading processes have also been observed in the context of adoption of emerging fashion trends within distinct age groups, the successive firing of thousands of neurons in the brain in response to stimuli, and plummeting stock prices in global financial markets in response to a natural disaster.  

Cascades are generally observed by simply recording the time when a specific website first mentioned a cascading news item, or when an infected person first showed symptoms of a disease. On the other hand, the underlying network topologies may be unknown and dynamic, with the link structure varying over time. Unveiling such 
dynamic network topologies is crucial for several reasons. Viral web advertising can be more effective if a small set of influential early adopters are identified through the link structure, while knowledge of the structure of hidden needle-sharing networks among communities of injecting drug users can aid formulation of policies for curbing contagious diseases. Other examples include assessment of the reliability of heavily interconnected systems like power grids, or risk exposure among investment banks in a highly inter-dependent global economy. In general, knowledge of topologies that facilitate diffusion of network processes leads to useful insights about the behavior of complex systems.

Network contagions arise due to \textit{causal} interactions between nodes e.g., blogs, or disease-susceptible individuals. \textit{Structural equation models} (SEMs) 
effectively capture such causal relationships, that are seldom revealed by symmetric correlations; see e.g.,~\cite{kaplan_book,pearl_book}. Widely applied in psychometrics~\cite{muthen}, and sociometrics~\cite{goldberger}, SEMs have also been adopted for gene network inference~\cite{juan1,logsdon}, and brain connectivity studies~\cite{mcintosh}. Recently, dynamic SEMs have been advocated for tracking slowly-varying sparse social network topologies from cascade data~\cite{baingana_jstsp}.

Network topologies may sometimes ``jump'' between a finite number of discrete states, as manifested by sudden changes in cascade behavior. For example, an e-mail network may switch topologies from predominantly work-based connections during the week, to friend-based connections over the weekend. Connection dynamics between bloggers may switch suddenly at the peak of sports events (e.g., ``Superbowl''), or presidential elections. In such settings, contemporary approaches assuming that network dynamics arise as a result of slow topology variations may yield unpredictable results. The present paper capitalizes on this prior knowledge, and puts forth a novel \emph{switched} dynamic SEM to account for propagation of information cascades in such scenarios. The novel approach builds upon the dynamic SEM advocated in~\cite{baingana_jstsp}, where it is tacitly assumed that node infection times depend on both the switching topologies and exogenous influences e.g., external information sources,  or prior predisposition to certain cascades. 

The present paper draws connections to identification of hybrid systems, whose behavior is driven by interaction between continuous and discrete dynamics; see e.g.,~\cite{paoletti} and references therein. Switched linear models have emerged as a useful framework to capture piecewise linear input-output relations in control systems~\cite{bako,roll,vidal}. The merits of these well-grounded approaches are broadened here to temporal network inference from state-driven cascade dynamics. Although the evolution of the unknown network state sequence may be controlled by structured hidden dynamics (e.g., \emph{hidden Markov models}), this work advocates a more general framework in which such prior knowledge is not assumed.

Network inference from temporal traces of infection events has recently emerged as an active research area. A number of approaches put forth probabilistic models for information diffusion, and leverage maximum likelihood estimation (MLE) to infer both static and dynamic edge weights as pairwise transmission weights between nodes~\cite{rodriguez2010,rodriguez2011,meyers2013}. Sparse SEMs are leveraged to capture exogenous inputs in inference of dynamic social networks in~\cite{baingana_jstsp}, and static causal links in gene regulatory networks~\cite{juan1}. Network Granger causality with group sparsity is advocated for inference of causal networks with inherent grouping structure in~\cite{basu}, while causal influences are inferred by modeling historical network events as multidimensional Hawkes processes in~\cite{zhou2013}.

Within the context of prior works on dynamic network inference, the contributions of the present paper are three-fold. First, a novel switched dynamic SEM that captures sudden topology changes within a finite state-space is put forth (Section~\ref{sec:model}). Second, identifiability results for the proposed switched model are established under reasonable assumptions, building upon prior results for static causal networks in~\cite{juan0} (Section~\ref{sec:ident}). Finally, an efficient sparsity-promoting proximal-gradient (PG) algorithm is developed (Sections~\ref{sec:splse} and~\ref{sec:algorithms}) to jointly track the evolving state sequence, and the unknown network topologies. Numerical tests on both synthetic and real cascades in Section~\ref{sec:experiments} corroborate the efficacy of the novel approach. Interestingly, experiments on real-world web cascade data exemplify that media influence is dominated by major news outlets (e.g., \emph{cnn.com} and \emph{bbc.com}), as well as well-known web-based news aggregators (e.g., \emph{news.yahoo.com} and \emph{news.google.com}).
\begin{figure*}[!tb]
\centering
\includegraphics[width=0.8\textwidth]{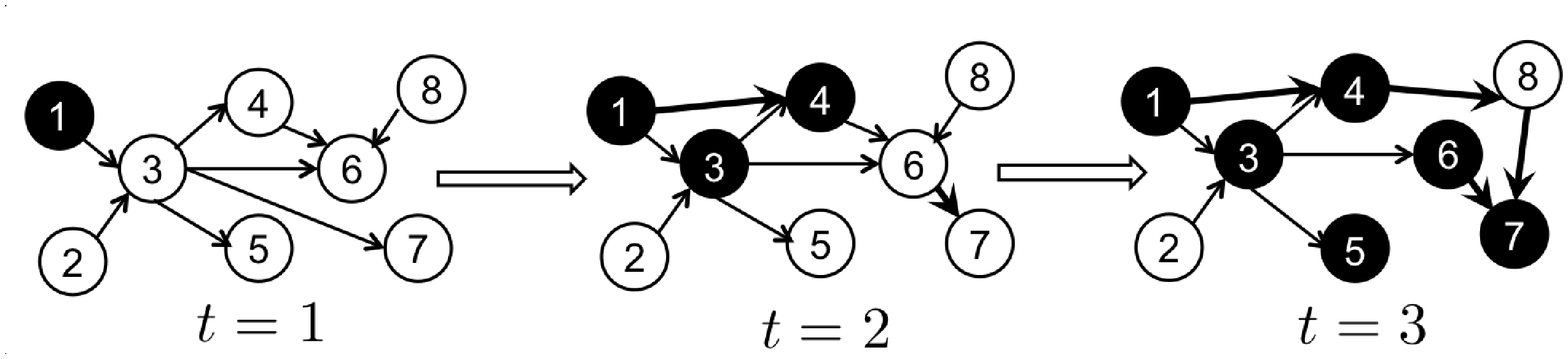}
\caption{A single cascade propagating over a time-varying network, with ``infected'' nodes depicted in black. Some edges disappear over consecutive time intervals as new ones emerge (thick arrows). Time-varying changes to the network topology affect the progression of the cascade over the network.}
\label{fig:fig_sem_illust0}
\end{figure*}

\noindent\textit{Notation}. Bold uppercase (lowercase) letters will denote matrices (column vectors), while operators $(\cdot)^{\top}$, $\lambda_{\max}(\cdot)$, and $\textrm{diag}(\cdot)$ will stand for matrix transposition, maximum eigenvalue, and diagonal matrix, respectively. The identity matrix will be represented by $\I$, while $\mathbf{0}$ will denote the matrix of all zeros, and their dimensions will be clear in context. Finally, the $\ell_p$ and Frobenius norms will be denoted by $\|\cdot\|_p$, and $\|\cdot\|_F$, respectively. 

\section{Model and Problem Statement}
\label{sec:model}
Consider a dynamic network with $N$ nodes observed over time intervals $t = 1, \dots, T$, captured by a graph whose topology may arbitrarily switch between $S$ discrete states. Suppose the network topology during interval $t$ is described by an unknown, weighted adjacency matrix $\A^{\sigma(t)} \in \mathbb{R}^{N \times N}$, with $\sigma(t) \in \mathcal{S} := \{1,\dots,S\}$. Entry $(i,j)$ of $\A^{\sigma(t)}$ (henceforth denoted by $a_{ij}^{\sigma(t)}$) is nonzero only if a directed edge connects nodes $i$ and $j$ (pointing from $j$ to $i$) during interval $t$. In general $a_{ij}^{\sigma(t)}\neq a_{ji}^{\sigma(t)}$, i.e., matrix $\A^{\sigma(t)}$ is generally non-symmetric, which is suitable to model directed networks. The model tacitly assumes that the network topology remains fixed during any given time interval $t$, but can change across time intervals.

Suppose $C$ contagions propagating over the network are observed, and the time of first infection per node $i$ by contagion $c$ is denoted by $y_{ic}^t \in \mathbb{R}$. In online media, $y_{ic}^t$ can be obtained by recording the timestamp when blog $i$ first mentioned news item $c$. If a node remains uninfected during $t$, $y_{ic}^t$ is set to an arbitrarily large number. Assume that the susceptibility $x_{ic}$ of node $i$ to external (non-topological) infection by contagion $c$ is known and time invariant. In the web context,  $x_{ic}$ can be set to the search engine rank of website $i$ with respect to (w.r.t.) keywords associated with $c$. 
\begin{figure}[!tb]
\centering
\includegraphics[width=0.3\textwidth]{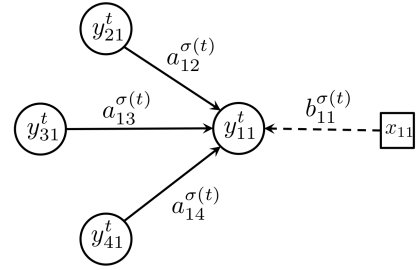}
\caption{The cascade infection time for node $1$ during interval $t$ is a linear combination of 
cascade infection times of its single-hop neighbors, and the exogenous influence
$x_{11}$. Unknown edge weights are captured through the 
coefficients $a_{12}^{\sigma(t)}, a_{13}^{\sigma(t)}$ and $a_{14}^{\sigma(t)}$.}
\label{fig:fig_sem_illust1}
\end{figure}

The infection time of node $i$ during interval $t$ is modeled according to the following \emph{switched} dynamic structural equation model (SEM)
\begin{equation}
\label{eq:mps1}
y_{ic}^t = \sum\limits_{j \ne i} a_{ij}^{\sigma(t)} y_{jc}^t + b_{ii}^{\sigma(t)} x_{ic} + e_{ic}^t
\end{equation}
where  $b_{ii}^{\sigma(t)}$ captures the state-dependent level of influence of external sources, and $e_{ic}^t$ accounts for measurement errors and unmodeled dynamics. It follows from \eqref{eq:mps1} that if $a_{ij}^{\sigma(t)}\neq 0$, then $y_{ic}^t $ is affected by the values of $\{y_{jc}^t\}_{j \neq i}$ (see Figure~\ref{fig:fig_sem_illust1}). With diagonal $\B^{\sigma(t)}:=\textrm{diag}(b_{11}^{\sigma(t)},\ldots,b_{NN}^{\sigma(t)})$, collecting observations for the entire network and all $C$ contagions yields the dynamic matrix SEM
\begin{equation}
\label{eq:mps3}
\Y_t = \A^{\sigma(t)} \Y_t + \B^{\sigma(t)} \X + \E_t
\end{equation}
where $\Y_t := \left[  y_{ic}^t\right]$, $\X := \left[ x_{ic}  \right]$, and $\E_t := \left[ e_{ic}^t\right] $ are all $N \times C$ matrices. A single network topology $\A^{\sigma(t)}$
is adopted for all contagions, which is suitable e.g., when information cascades are formed around a common meme or trending (news) topic over the Internet. Matrix $\A^{\sigma(t)}$ has an all-zero diagonal to ensure that the underlying network is free of self loops, i.e., $a_{ii}^{\sigma(t)} = 0 \;\; \forall i, \forall t$.

\noindent\textbf{Problem statement.} 
Given the sequence $\{ \Y_t \}_{t=1}^T$ and $\X$ adhering to \eqref{eq:mps3}, the goal is to identify the unknown state matrices $ \{\A^s, \B^s\}_{s=1}^S $, and the switching sequence $\{\sigma(t) \in \mathcal{S}\}_{t=1}^T$.

\section{Model Identifiability}
\label{sec:ident}
This section explores the conditions under which one can uniquely recover the unknown state matrices $ \left\lbrace \A^{s}, \B^{s} \right\rbrace_{s=1}^S $. First, \eqref{eq:mps3} can be written as 
\begin{equation}
\label{eq:id1}
\Y_t = \left( \sum\limits_{s=1}^S \chi_{ts} \A^s \right) \Y_t + \left( \sum\limits_{s=1}^S  \chi_{ts} \B^s \right) \X + \E_t
\end{equation}
where the binary indicator variable $\chi_{ts} = 1$ if $\sigma(t) = s$, otherwise $\chi_{ts} = 0$. Note that \eqref{eq:id1} is generally underdetermined when $S$ is unknown, and it may not be possible to uniquely identify $\{ \A^s, \B^s, \{ \chi_{ts} \}_{t=1}^T \}_{s=1}^S$. Even in the worst-case scenario where $S=T$, complete identification of all states is impossible if there exists $t$ and $t'$ ($t \neq t'$) such that $\sigma(t) = \sigma(t')$. In order to establish model identifiability, it will be assumed that the following hold: 

\noindent \textbf{as1.} The dynamic SEM in \eqref{eq:mps3} is noise-free ($ \E_t = \mathbf{0} $); that is,
\begin{equation}
\label{eq:id0}
\Y_t = \A^{\sigma(t)} \Y_t + \B^{\sigma(t)} \X.
\end{equation}

\noindent \textbf{as2.} All cascades $ \left\lbrace \Y_t \right\rbrace_{t=1}^T $ are generated by some pair $ \{ \A^s, \B^s \} $, where $ s \in \left\lbrace 1, \dots, S \right\rbrace $, and $S$ is known. This is a \emph{realizability assumption}, which guarantees the existence of a submodel responsible for the observed cascades.

\noindent \textbf{as3.} No two states can be active during a given time interval, i.e., $ \| \Y_t - \A^s \Y_t - \B^s \X \|_F =  \| \Y_t - \A^{s'} \Y_t - \B^{s'} \X \|_F = 0$ implies that $s = s'$.

\noindent Under (as1)-(as3), the following proposition holds.

\begin{proposition}
\label{prop:ml}
Suppose data matrices $\Y_t$ and $\X$ adhere to \eqref{eq:id0} with $a_{ii}^{\sigma(t)}=0 $, $b_{ii}^{\sigma(t)} \neq 0, \;\; \forall \; i, t$, and   $b_{ij}^{\sigma(t)} = 0 \;\; \forall \; i \neq j, t$. If $N \leq C$ and $\X$ has full row rank, then $\A^{\sigma(t)}$ and $\B^{\sigma(t)}$ are uniquely expressible in terms of $\X$ and $\Y_t$ as $\B^{\sigma(t)} = \left( \text{Diag}\left[ \left( \Y_t \X^{\dagger} \right)  \right]  \right)^{-1}$, and  $\A^{\sigma(t)} = \I - \B^{\sigma(t)} \left( \Y_t \X^{\dagger} \right)^{-1}$.
 
\noindent If $p_s := \text{Pr} \left( \chi_{ts} = 1 \right) > 0 \; \forall s$ denotes the activation probability of state $s$, with $\sum_{s=1}^S p_s = 1$, and (as1)-(as3) hold, then  $\left\lbrace \A^s, \B^s \right\rbrace_{s=1}^S$ can be uniquely identified with probability one
 as $T \rightarrow \infty$.
\end{proposition}
 
\noindent Proposition~\ref{prop:ml} establishes a two-step identifiability result for the dynamic SEM model in \eqref{eq:id0}. First, it establishes that if the exogenous data matrix $\X$ is sufficiently rich (i.e., $\X$ has full row rank), then the per-interval network topology captured by $\B^{\sigma(t)}$ can be uniquely identified per $t$. Second, if the state activation probabilities are all strictly positive, then the $S$ switching network topologies will all be recovered as $T \rightarrow \infty$. In fact, if cascade data are acquired in an infinitely streaming fashion ($T = \infty$), then one is guaranteed to uniquely recover all states.

\noindent \textbf{Proof of Proposition~\ref{prop:ml}:}  Equation~\eqref{eq:id0} can be written as $(\I - \A^{\sigma(t)}) \Y_t = \B^{\sigma(t)} \X$, which implies that $\text{rank}\left( (\I - \A^{\sigma(t)}) \Y_t  \right) = \text{rank} \left(  \B^{\sigma(t)} \X \right)$. With $\X$ and $\B^{\sigma(t)}$ both having full row rank (recall that 
$b^{\sigma(t)}_{ii} > 0$), it follows that  $(\I - \A^{\sigma(t)}) \Y_t$ has full row rank, and $\I - \A^{\sigma(t)}$ is invertible. Consequently, the linear system of equations $\Y_t = \Phib \X$ is solved by $\Phib^{\ast} = \left( \I - \A^{\sigma(t)} \right)^{-1}\B^{\sigma(t)}$.

On the other hand, note that $\Y_t = \Phib \X$ admits the solution 
\begin{equation}
\label{eq:id2}
\Phib^{\ast} = \Y_t \X^{\dagger}
\end{equation}
which is unique since $\X$ is full row rank. Since $a_{ii}^{\sigma(t)} = 0$ and $\left( \B^{\sigma(t)} \right)^{-1}$ is a diagonal matrix, the diagonal entries of $(\Phib^{\ast})^{-1} = \left( \B^{\sigma(t)} \right)^{-1} \left( \I - \A^{\sigma(t)} \right)$ coincide with those of $\left( \B^{\sigma(t)} \right)^{-1}$; hence, $\B^{\sigma(t)} = \left( \text{Diag}\left[ \left( \Phib^{\ast}  \right)^{-1}  \right]\right)^{-1} $, which leads to
\begin{equation}
\label{eq:id3}
\B^{\sigma(t)} = \left( \text{Diag}\left[ \left( \Y_t \X^{\dagger}  \right)^{-1}  \right]\right)^{-1}
\end{equation}
upon substitution of~\eqref{eq:id2}. Furthermore, note that $\B^{\sigma(t)} \left( \Y_t \X^{\dagger} \right)^{-1} = \I - \A^{\sigma(t)}$, and thus
\begin{equation}
\label{eq:id4}
\A^{\sigma(t)} = \I - \left[ \text{Diag} \left( \left(  \Y_t \X^{\dagger} \right)^{-1} \right) \right]^{-1}\left( \Y_t \X^{\dagger} \right)^{-1}
\end{equation}
which concludes the first part of the proof.

Since $\A^{\sigma(t)} =  \sum\limits_{s=1}^S \chi_{ts} \A^s $ and $\B^{\sigma(t)} =  \sum\limits_{s=1}^S  \chi_{ts} \B^s $, the results from the unique solutions~\eqref{eq:id3} and~\eqref{eq:id4}
coincide with a specific pair in the set $\left \lbrace \left(\A^s, \B^s \right) \right \rbrace_{s=1}^S$ per $t$. Intuitively, complete recovery of all
state matrices is tantamount to identification of $S$ unique pairs $\left( \A^s, \B^s \right)$ as more data are sequentially acquired. In order to guarantee identifiability of all states in the long run, it suffices to prove that 
the probability of activation of any state at least once tends to $1$ as $T \rightarrow \infty$. Since $\chi_{ts}$ are Bernoulli random variables with $p_s = \text{Pr} ( \chi_{ts} = 1  )$ per $t$, the number of times that state $s$ is
activated over $T$ time intervals follows a binomial distribution. Letting $T_s := \sum\limits_{t=1}^T \chi_{ts}$  denote 
the total number of activations of state $s$ over $t = 1, \dots, T$, then 
\begin{eqnarray}
\label{eq:id5}
\nonumber
\text{Pr}(T_s \geq 1) & = & 1 - \text{Pr}(T_s = 0) \\
 &=& 1 - (1 - p_s)^T.
\end{eqnarray}
Since $\underset{T \rightarrow \infty}{\text{lim}} \; \text{Pr}(T_s \geq 1) = 1$ only if $p_s > 0$, it follows that all states can be uniquely identified with probability $1$ as $T \rightarrow \infty$ when $p_s > 0 \;\; \forall s$, which completes
the second part of the proof.

\subsection{Topology tracking by clustering when $\E_t \neq \mathbf{0}$}
\label{ssec:noisystates}
In general, even when $\X$ is full row rank, $\E_t \neq \mathbf{0}$ in order to compensate for measurement errors and unmodeled dynamics. Under noisy conditions, it will turn out that $\{ \A^s, \B^s \}_{s=1}^S$ can be interpreted as cluster centroids (cf.~\eqref{eq:id1}), and one can leverage traditional clustering approaches (e.g., k-means), coupled with the closed-form solutions in~\eqref{eq:id3} and~\eqref{eq:id4} to identify the unknown state matrices. Indeed, with $\b^{\sigma(t)} := \text{diag}\left( \B^{\sigma(t)} \right)$, and $\boldsymbol{\theta}^t := \text{vec}\left( \left[  \A^{\sigma(t)} \; \b^{\sigma(t)}\right]  \right)$, it follows readily that
\begin{equation}
\label{eq:id6}
\boldsymbol{\theta}^t = \sum\limits_{s=1}^S \chi_{ts} \boldsymbol{\theta}^s \;\; t = 1, \dots, T
\end{equation}
where $\boldsymbol{\theta}^s := \text{vec}\left( \left[ \A^s \; \b^s \right] \right)$, and $\b^s := \text{diag}(\B^s)$. 

Clearly, $\{ \boldsymbol{\theta}^s  \}_{s=1}^S$ in~\eqref{eq:id6} can be viewed as $S$ cluster centers, and $\chi_{ts}$ as unknown binary cluster-assignment variables. Consequently, the sequence $\{ \boldsymbol{\theta}^t \}_{t=1}^{T_{\text{cluster}}}$ can be directly computed per $t$ via~\eqref{eq:id3} and~\eqref{eq:id4}, where $T_{\text{cluster}} < T$ denotes the number of training samples. Identification of $\{ \chi_{ts} \}$  and $\{ \boldsymbol{\theta}^s \}$ can then be accomplished by batch clustering $\{ \boldsymbol{\theta}^t \}$ into $S$ clusters. The subsequent operational phase ($t > T_{\text{cluster}}$) then boils down to computing $\boldsymbol{\theta}^t$, followed by finding the centroid $\boldsymbol{\theta}^s$ to which it is closest in Euclidean distance, that is
\begin{equation}
\label{eq:id7}
\hat{s} = \underset{s}{\text{arg min}} \;\; \| \boldsymbol{\theta}^t - \boldsymbol{\theta}^s  \|_2
\end{equation}
and $\hat{\chi}_{ts} = 1$ if $s = \hat{s}$, otherwise $\hat{\chi}_{ts} = 0$. Algorithm~\ref{alg0} summarizes this cluster-based state identification scheme, with a clustering phase ($t \leq T_{\text{cluster}}$) and an operational phase ($t > T_{\text{cluster}}$). The sub-procedure $ \text{cluster}(\{ \boldsymbol{\theta}^t \}_{t=1}^{T_{\text{cluster}}}, S)$ calls an off-the-shelf clustering algorithm with the training set $\{ \boldsymbol{\theta}^t \}_{t=1}^{T_{\text{cluster}}}$ and the number of clusters $S$ as inputs.

\begin{remark}
\label{rem:soft_clustering}
\normalfont Algorithm~\ref{alg0} identifies the unknown state-dependent matrices by resorting to ``hard clustering,'' which entails deterministic assignment of $\boldsymbol{\theta}^t$ to one of the $S$ centroids. In principle, the algorithm can be readily modified to adopt ``soft clustering'' approaches, with probabilistic assignments to the cluster centroids. 
\end{remark}
\begin{algorithm}[t!]
    \caption{Topology identification by clustering}
\label{alg0}
\begin{algorithmic}[1]
   \REQUIRE  $\left\lbrace \Y_{t} \right\rbrace_{t=1}^T$, $\X$, $S$, $T_{\text{cluster}}   < T$
   \FOR{$t=1,\dots,T_{\text{cluster}}$} 
   \STATE $\b^{\sigma(t)} = \text{diag}[( \text{Diag}[( \Y_t \X^{\dagger})^{-1} ])^{-1}]$
   \STATE $\A^{\sigma(t)} = \I - [ \text{Diag} ( (  \Y_t \X^{\dagger} )^{-1} ) ]^{-1}( \Y_t \X^{\dagger} )^{-1}$
   \STATE $\boldsymbol{\theta}^t := \text{vec}\left( \left[  \A^{\sigma(t)} \; \b^{\sigma(t)}\right]  \right)$
   \ENDFOR
   \STATE $\{ \boldsymbol{\theta}^s, \{ \chi_{ts} \}_{t=1}^{{T_{\text{cluster}}}}\}_{s=1}^S = \text{cluster}(\{ \boldsymbol{\theta}^t \}_{t=1}^{T_{\text{cluster}}}, S)$
   \STATE Set $\hat{\sigma}(t) = s$ if $\chi_{ts} = s$ for $t = 1,\dots, T_{\text{cluster}}$
   \STATE Extract $\{ \A^s, \B^s\}_{s=1}^S$ from $\{ \boldsymbol{\theta}^s \}_{s=1}^S$
   \FOR{$t > T_{\text{cluster}}$}
   \STATE Compute $\boldsymbol{\theta}^t $ using lines $(2) - (4)$
   \STATE $\hat{\sigma}(t) = \underset{s}{\text{arg min}} \;\; \| \boldsymbol{\theta}^t - \boldsymbol{\theta}^s  \|_2$
   \ENDFOR
   \RETURN $\{ \A^s, \B^{s} \}_{s=1}^S, \{ \hat{\sigma}(t) \}_{t=1}^T$
\end{algorithmic}
\end{algorithm}
 %
\section{Exploiting edge sparsity}
\label{sec:splse}
The fundamental premise established by Proposition~\ref{prop:ml} is that $\{ \A^s, \B^s \}_{s=1}^S$ are uniquely identifiable provided $ \X $ is sufficiently rich. However, requiring that $ \X $ has full row rank is a restrictive condition, tantamount to requiring that at least as many cascades are observed as the number of nodes ($C \geq N$). This is especially prohibitive in large-scale networks such as the world-wide web, with billions of nodes. It is therefore of interest to measure as few cascades as possible while ensuring that $\{ \A^s, \B^s \}_{s=1}^S$ are uniquely recovered. To this end, one is motivated to leverage prior knowledge about the unknowns in order to markedly reduce the amount of data required to guarantee model identifiability. For example, each node in the network is connected only to a small number of nodes out of the $N-1$ possible connections. As a result, most practical networks exhibit edge sparsity, a property that can be exploited to ensure that the rank condition on $\X$ can be relaxed, as shown in the sequel. In addition to (as1)-(as3), consider the following.

\noindent \textbf{as4.} Each row $\a_n^s$ of matrix $\A^s$ has at most $K$ nonzero entries; i.e., $\| \a_n^s \|_0 \leq K \;\; \forall n,s$. 

\noindent \textbf{as5.} The nonzero entries of $\A^s$ for all $s=1,\dots,S$ are drawn from a continuous distribution.

\noindent \textbf{as6.} The Kruskal rank of $\X^{\top}$ satisifies $\text{kr}\left( \X^{\top} \right) \geq 2K + 1$, where $\text{kr} \left( \mathbf{Z} \right)$ is defined as the maximum number $k$ such that any combination of $k$ columns of $\mathbf{Z}$ constitute a full column rank submatrix. 

\begin{proposition}
\label{prop:ml2}
Suppose data matrices $\Y_t$ and $\X$ adhere to \eqref{eq:id0} with $a_{ii}^{\sigma(t)}=0 $, $b_{ii}^{\sigma(t)} \neq 0, \;\; \forall \; i, t$, and   $b_{ij}^{\sigma(t)} = 0 \;\; \forall \; i \neq j, t$. If (as1)-(as6) hold, and $p_s > 0 \;\; \forall s$, then  $\left\lbrace \A^s, \B^s \right\rbrace_{s=1}^S$ can be uniquely identified with probability one as $T \rightarrow \infty$.
\end{proposition}

The proof of Proposition~\ref{prop:ml2} is rather involved, and is deferred to Appendix~\ref{app:proof1}. Unlike the matrix rank which only requires existence of a subset of linearly independent columns, the Kruskal rank requires that every possible combination of a given number of columns be linearly independent. Moreover, computing $\text{kr} \left( \X^{\top} \right)$ is markedly more challenging, as it entails a combinatorial search over the rows of $\X$. Admittedly, requiring $\text{kr}\left( \X^{\top} \right) \geq 2K+1$ is more restrictive than $\text{rank}\left( \X^{\top} \right) \geq 2K+1$. Nevertheless, in settings where $2K + 1 \ll N$, (as6) may be satisfied even if $\text{rank} \left( \X \right) < N$. In such cases, Proposition~\ref{prop:ml2} asserts that one can uniquely identify $\{ \A^s, \B^s  \}_{s=1}^S$ even when $C \leq N$.

\subsection{Sparsity-promoting estimator}
\label{ssec:spe}
The remainder of the present paper leverages inherent edge sparsity to develop an efficient algorithm to track switching network topologies from noisy cascade traces. Assuming $S$ is known a priori, one is motivated to solve the following regularized LS batch estimator
\begin{eqnarray}
\label{eq:swtch2}
\nonumber \text{(P0)} \;\;
\underset{\substack{ \{\A^s, \B^s\}_{s=1}^{S} \\ \{ \{ \chi_{ts} \}_{s=1}^S \}_{t=1}^T}}{\text{arg min}} & &
\frac{1}{2}\sum\limits_{t=1}^{T} \sum\limits_{s=1}^{S} \chi_{ts} 
\| \Y_t - \A^s \Y_t - \B^s\X  \|_F^2 \\ 
\nonumber
& & + \sum\limits_{s=1}^{S} \lambda_s \| \A^s \|_1 \\
\nonumber
\text{s. to} & & \sum\limits_{s=1}^{S} \chi_{ts} = 1 \; \forall t, \;\;
\chi_{ts} \in \{0,1\} \; \forall s, t \\
& &  a_{ii}^s = 0, \; b_{ij}^s = 0, \; \forall s,  i \ne j
\end{eqnarray}
where the constraint $\sum_{s=1}^{S} \chi_{ts} = 1$ enforces a realizability condition, ensuring that only one state can account for the system behavior at any time. With $\| \A^s \|_1 := \sum_{ij} |a^s_{ij}| $, the regularization term promotes edge sparsity that is inherent to most real networks. The sparsity level of $\hat{\A}^s$ is controlled by  $\lambda_s > 0$. Absence of a self-loop at node $i$ is enforced by the constraint $a_{ii}^s = 0$, while having $b_{ij}^s = 0, \; \forall  i \ne j $, ensures that $\hat{\B}^s$ is diagonal as in \eqref{eq:mps3}.

Note that (P0) is an NP-hard mixed integer program that is unsuitable for large-scale and potentially real-time operation. Moreover, entries of per-state matrices $\{\A^s, \B^s \}$ may evolve slowly over reasonably long observation periods, motivating algorithms that not only track the switching sequence, but also the slow drifts occurring in per-state network topologies. 

Suppose data are sequentially acquired, rendering batch estimators impractical. If it is assumed that instantaneous and past estimates of $\{ \{ \hat{\chi}_{\tau s} \}_{s=1}^S \}_{\tau=1}^{t}$ are known during interval $t$, (P0) decouples over $\mathcal{S}$, and is tantamount to solving the following subproblem per $t$ and $s$
\begin{eqnarray}
\label{eq:swtch2b}
\nonumber \text{(P1)} \;\;\;\;
\underset{ \A^s, \B^s }{\text{arg min}} & &
\frac{1}{2}\sum\limits_{\tau=1}^{t} \hat{\chi}_{\tau s} 
\| \Y_{\tau} - \A^s \Y_{\tau} - \B^s\X  \|_F^2 \\ 
\nonumber
& & + \lambda_s \| \A^s \|_1 \\
\text{s. to} 
& &  a_{ii}^s = 0, \; b_{ij}^s = 0, \; \forall  i \ne j.
\end{eqnarray}
Note that the LS term in the penalized cost only aggregates residuals when $\hat{\chi}_{\tau s} = 1 \; (\hat{\sigma}(t) = s)$. In principle, during interval $t$, (P1) updates the 
estimates of $\A^s$ and $\B^s$ only if $\Y_t$ has been generated by submodel $s$. Critical to efficiently tracking the hidden network topologies is the need for a reliable approach to estimate $\chi_{\tau s}$. Before developing an efficient tracking algorithm that will be suitable to solve (P1), the rest of this section puts  forth criteria for estimating $\sigma(t)$.

\noindent\textbf{Estimation of $\boldsymbol{\sigma(t)}$.} Given the most recent estimates $\{ \hat{\A}^s, \hat{\B}^s  \}_{s=1}^S$ prior to acquisition of $\Y_t$, one can estimate $\sigma(t)$ by minimizing the a priori error over $\mathcal{S}$ i.e.,
\begin{equation}
\label{eq:swtch3}
\hat{\sigma}(t) = \arg \underset{s \in \mathcal{S}}{\min} \;\;
\| \Y_t - \hat{\A}^s \Y_t - \hat{\B}^s \X \|_F
\end{equation}
followed by solving (P1) with $s = \hat{\sigma}(t)$. If network dynamics arise from switching between static or slowly-varying per-state topologies,~\eqref{eq:swtch3} yields a reliable estimate of the most likely state sequence over time.

Alternatively, one may resort to a criterion that depends on minimizing the a posteriori error. This entails first solving (P1) for all states $s \in \mathcal{S}$ upon acquisition of $\Y^t$, and then selecting $s = \hat{\sigma}(t)$ so that
\begin{equation}
\label{eq:swtch3p}
\hat{\sigma}(t) = \arg \underset{s \in \mathcal{S}}{\min} \;\;
\| \Y_t - \hat{\A}^{(s|\sigma(t) = s)} \Y_t - \hat{\B}^{(s|\sigma(t) = s)} \X \|_F
\end{equation}
where $\hat{\A}^{(s|\sigma(t) = s)}$ (resp. $\hat{\B}^{(s|\sigma(t) = s)}$) denotes the estimate of $\A^s$ (resp. $\B^s$) given that $\sigma(t) = s$. In the context of big data and online streaming, \eqref{eq:swtch3p} is prohibitive, and \eqref{eq:swtch3} is more desirable for its markedly lower computational overhead. The tracking algorithm developed next adopts~\eqref{eq:swtch3} for state sequence estimation.

\section{Topology Tracking Algorithm}
\label{sec:algorithms}
In order to solve (P1), this section resorts to proximal gradient (PG) approaches, which have been popularized for $\ell_1$-norm regularized problems, through the class of iterative shrinkage-thresholding algorithms (ISTA); see e.g.,~\cite{daubechies} and~\cite{boyd2}. Unlike off-the-shelf interior point methods, ISTA is computationally simple, with iterations entailing matrix-vector multiplications, followed by a soft-thresholding operation~\cite[p. 93]{hastie_book}. Motivated by its well-documented merits, an ISTA algorithm is developed in the sequel for recursively solving (P1) per $t$. Memory storage and computational costs incurred by the algorithm per acquired sample $\Y_t$ do not grow with $t$.  

\noindent\textbf{Solving (P1) for a single time interval $\mathbf{t}$.} 
Introducing the optimization variable $\V^s:=[\A^s\:\:\B^s]$, it follows that the gradient of $f(\V^s):=(1/2) \sum_{\tau=1}^{t} \hat{\chi}_{\tau s}   \| \Y_\tau - \A^s \Y_\tau - \B^s \X  \|_F^2$ is Lipschitz continuous, meaning there exists a constant $L_f$ so that $\|\nabla f(\V_1^s)-\nabla f(\V_2^s)\|\leq L_f\|\V_1^s-\V_2^s\|$, $\forall\:\V_1^s,\:\V_2^s$ in the domain of $f$. Instead of directly optimizing the cost in (P1), PG algorithms minimize a sequence of overestimators evaluated at the current iterate, or a linear combination of the two previous iterates.

Letting $k=1,2,\ldots$ denote the iteration index, and $g(\V^s):=\lambda_s\|\A^s\|_{1}$, PG algorithms iterate
\begin{align}\label{eq:pg_iterates}
\V^s[k]{}:={}&\arg\min_{\V}
\left\{\frac{L_f}{2}\|\V-\G(\V^s[k-1])\|_F^2+g(\V)\right\}
\end{align}
where $\G(\V^s[k-1]){:=}\V^s[k-1]-(1/L_f)\nabla f(\V^s[k-1])$ corresponds to a gradient-descent step taken from $\V^s[k-1]$, with step-size equal to $1/L_f$.  The optimization problem \eqref{eq:pg_iterates} is denoted by $\textrm{prox}_{g/L_f}(\G(\V^s[k-1]))$, and is known as the proximal operator of the function $g/L_f$ evaluated at $\G(\V^s[k-1])$. 
With $\G^s[k-1]:=\G(\V^s[k-1])$ for convenience, the PG iterations can be compactly rewritten as $\V[k]=\textrm{prox}_{g/L_f}(\G^s[k-1])$.

The success of PG algorithms hinges upon efficient evaluation of the proximal operator (cf. \eqref{eq:pg_iterates}). Focusing on (P1), note that \eqref{eq:pg_iterates}
decomposes into
\begin{align}
\A^s[k]{}:={}&\arg\min_{\A^s}
\left\{\frac{L_f}{2}\|\A^s-\G_A^s[k-1]\|_F^2+\lambda_s\|\A^s\|_1\right\}\nonumber\\
\label{eq:pg_iterates_A}{}={}&\text{soft}_{\lambda_s/L_f}(\G_A^s[k-1])\\
\label{eq:pg_iterates_B}\B^s[k]{}:={}&\arg\min_{\B^s}
\left\{\|\B^s-\G_B^s[k-1]\|_F^2\right\}=\G_B^s[k-1]
\end{align}
subject to the constraints in (P1) which so far have been left implicit, and $\G^s {:=} [\G_A^s\: \G_B^s]$. Letting $\text{soft}_{\mu}(\M)$ with $(i,j)$-th entry given by $\textrm{sign}(m_{ij})\max(|m_{ij}|-\mu,0)$ denote the soft-thresholding operator, it follows that $\textrm{prox}_{\lambda_s\|\cdot\|_1/L_f}(\cdot)=\text{soft}_{\lambda_s/L_f}(\cdot)$, see e.g.,~\cite{daubechies,hastie_book}. Since there is no regularization on $\B^s$, \eqref{eq:pg_iterates_B} boils-down to a simple gradient-descent step.

Specification of $\G_A^s$ and $\G_B^s$ only requires expressions for the gradient of  $f(\V^s)$ with respect to $\A^s$ and $\B^s$. Note that by incorporating the constraints $a^s_{ii} = 0$ and $b^s_{ij} = 0, \; \forall  j \ne i$, $i=1,\ldots N,$ one can express $f(\V^s)$ as
\begin{align}\label{eq:f_rowwise}
f(\V^s):=\frac{1}{2} \sum_{\tau=1}^{t}\sum_{i=1}^N \hat{\chi}_{\tau s}   \| \y_{i,\tau}^\top - (\a_{-i}^s)^\top\Y_{-i,\tau} - b_{ii}^s\x_i^\top  \|_2^2
\end{align}
where $\y_{i,\tau}^\top$ and $\x_i^\top$ denote the $i$-th row of $\Y_\tau$ and $\X$, respectively; while $(\a_{-i}^s)^\top$ denotes the $1\times (N-1)$ vector obtained by removing entry $i$ from the $i$-th row of $\A^s$, and likewise  $\Y_{-i,\tau}$ is the $(N-1)\times C$ matrix obtained by removing row $i$ from $\Y_{\tau}$. It is apparent from \eqref{eq:f_rowwise} that $f(\V^s)$ is separable across the trimmed row vectors $(\a_{-i}^s)^\top$, and the diagonal entries $b_{ii}^s$, $i=1,\ldots, N$. The sought gradients are 
\begin{align}
\label{eq:nabla_f_ai}\nabla_{\a_{-i}^s} f(\V^s)={}&\bm{\Omega}_{-i,t}^s\a_{-i}^s+\bar{\Y}^s_{-i,t}\x_i b_{ii}^s-\bm\omega_{-i,t}^s\\
\label{eq:nabla_f_bii}\nabla_{b_{ii}^s} f(\V^s)={}&(\a_{-i}^s)^\top\bar{\Y}^s_{-i,t}\x_i+ \alpha_t^s b_{ii}^s\|\x_i\|_2^2-
(\bar{\y}^s_{i,t})^\top\x_i
\end{align}
where $(\bar{\y}^s_{i,t})^\top$ denotes the $i$-th row of $\bar{\Y}^s_t:=\sum_{\tau=1}^{t} \hat{\chi}_{\tau s}\Y_{\tau}$, $\bar{\Y}^s_{-i,t}:=
\sum_{\tau=1}^{t}\hat{\chi}_{\tau s}\Y_{-i,\tau}$, and $\alpha_t^s := \sum_{\tau=1}^t \hat{\chi}_{\tau s}$. Similarly, $\bm\omega_{-i,t}^s:=\sum_{\tau=1}^{t}\hat{\chi}_{\tau s}\Y_{-i,\tau}\y_{i,\tau}$ and $\bm{\Omega}_{-i,t}^s$ is obtained by removing the $i$-th row and $i$-th column from $\bm{\Omega}^s_t:=\sum_{\tau=1}^{t}\hat{\chi}_{\tau s}\Y_{\tau} \Y_{\tau}^{\top}$. From \eqref{eq:pg_iterates_A}-\eqref{eq:pg_iterates_B} and  \eqref{eq:nabla_f_ai}-\eqref{eq:nabla_f_bii}, the parallel ISTA iterations
\begin{align}
\label{eq:nabla_f_ai_eval}\nabla_{\a_{-i}^s} f[k]={}&\bm{\Omega}_{-i,t}^s\a_{-i}^s[k]+\bar{\Y}^s_{-i,t}\x_i b_{ii}^s[k]-\bm\omega_{-i,t}^s\\
\label{eq:nabla_f_bii_eval}\nabla_{b_{ii}^s}f[k]={}&(\a_{-i}^s)^\top[k]\bar{\Y}^s_{-i,t}\x_i+ \alpha_t^s b_{ii}^s[k]\|\x_i\|_2^2-
(\bar{\y}^s_{i,t})^\top\x_i\\
\label{eq:pg_iters_ai}\hspace{-0.5cm}\a_{-i}^s[k+1]={}&\text{soft}_{\lambda_s/L_f}\left(\a_{-i}^s[k]-(1/L_f)\nabla_{\a_{-i}^s} f[k]\right)\\
\label{eq:pg_iters_bii} b_{ii}^s[k+1]={}& b_{ii}^s[k]-(1/L_f)\nabla_{b_{ii}^s}f[k]
\end{align}
are provably convergent to the globally optimal solution $\{\hat{\A}^s, \hat{\B}^s\}$ of (P1), as per general convergence results for PG methods~\cite{daubechies,boyd2}.

Note that iterations \eqref{eq:pg_iters_ai}-\eqref{eq:pg_iters_bii} incur a low computational overhead, involving at most matrix-vector multiplication complexity for the gradient evaluations. A final step entails zero-padding the updated $\a_{-i}^s[k]$ by setting
\begin{equation}
\label{eq:zero_pad}
(\a_i^s)^\top[k]=[a_{-i,1}^s[k]\ldots a_{-i,i-1}^s[k]\:0\:a_{-i,i}^s[k]\ldots a_{-i,N}^s[k]].
\end{equation}
The desired SEM parameter estimates are subsequently obtained as $\hat{\A}^{s}=[(\a_1^s)^\top[k],\ldots,(\a_N^s)^\top[k]]^\top$ and $\hat{\B}^{s}= \textrm{diag}(b_{11}^s[k],\ldots,b_{NN}^s[k])$, for $k$ large enough so that convergence has been attained.

\noindent\textbf{Solving (P1) over the entire time horizon.} 
Tracking the switching sequence and the state parameters entails sequentially alternating between two operations per datum arrival. First, $\hat{\sigma}(t)$ is estimated (via solving \eqref{eq:swtch3} or \eqref{eq:swtch3p}), and the corresponding values $\{ \hat{\chi}_{t s} \}_{s=1}^S$ are accordingly obtained. The iteration steps \eqref{eq:nabla_f_ai_eval}-\eqref{eq:pg_iters_bii} are then run until convergence is attained. Note that these iterations only depend on past data through recursively updated moving averages, that is,
\begin{align}
\label{eq:ta7a}
\bm{\Omega}^s_t = \bm{\Omega}^s_{t-1} + \hat{\chi}_{t s}\Y_t \Y_t^\top,\quad
\bar{\Y}^s_{t} = \bar{\Y}^s_{t-1} + \hat{\chi}_{ts}\Y_t.
\end{align}
Similar recursive expressions can be readily derived for $\bar{\Y}^s_{-i,t}, \alpha_t^s$, and $\boldsymbol{\omega}^s_{-i,t}$. The complexity in evaluating the Gram matrix $\Y_t\Y_t^{\top}$ dominates the per-iteration computational cost of the algorithm. The recursive updates in \eqref{eq:ta7a} are conducted only for a single state $s = \hat{\sigma}(t)$ per interval $t$, with the remainder $\{ \bm{\Omega}^s_t, \bar{\Y}^s_{t} \}_{s \in \mathcal{S} \setminus \hat{\sigma}(t)}$ set to $\{ \bm{\Omega}^s_{t-1}, \bar{\Y}^s_{t-1} \}_{s \in \mathcal{S} \setminus \hat{\sigma}(t)}$. Similarly, iterations \eqref{eq:nabla_f_ai_eval}-\eqref{eq:pg_iters_bii} are run only for $s = \hat{\sigma}(t)$, while $\{ \hat{\A}^s, \hat{\B}^s  \}_{s \in \mathcal{S} \setminus \hat{\sigma}(t)}$ are not updated during interval $t$. Furthermore, the need to recompute $L_f$ per $t$ can be circumvented by selecting 
an appropriate step-size for \eqref{eq:pg_iters_ai}-\eqref{eq:pg_iters_bii} by line search~\cite{boyd2}.

\begin{algorithm}[t!]
    \caption{Topology tracking algorithm}
\label{alg1}
\begin{algorithmic}[1]
   \REQUIRE  $\left\lbrace \Y_{t} \right\rbrace_{t=1}^T$,  $\X$, $S$, $\{ \lambda_s \}_{s=1}^S$
   \STATE Initialize $\{ \hat{\A}_0^s, \; \hat{\B}^s_0, \; \bar{\Y}_0^s = \mathbf{0}_{N \times C} \}_{s=1}^S$
   \FOR{$t = 1, \dots, T$}
   \STATE $\hat{\sigma}(t) = \text{arg min}_{s \in \mathcal{S}} \| \Y_t - \hat{\A}^s_{t-1}\Y_t - \hat{\B}^s_{t-1}\X  \|_F$
   \STATE Set $\hat{\chi}_{t \hat{\sigma}(t)} = 1$ and $\hat{\chi}_{t s} = 0$ for $s \in \mathcal{S} \setminus \hat{\sigma}(t)$
   \FOR{$s = 1, \dots, S$}
   \IF{$s = \hat{\sigma}(t)$}
   \STATE Update $\bm\Omega^{s}_t, \bar{\Y}^{s}_t, \alpha^s_t, L_f$
   \STATE Set $\A^s[0]=\hat{\A}^s_{t-1}$, $\B^s[0]=\hat{\B}^s_{t-1}$, $k=0$
   \WHILE {not converged}
   \FOR {$i=1 \dots N$ (in parallel)} 
   \STATE $\mathbf{z}^s[k] = \a_{-i}^s[k]{-}(1/L_f)\nabla_{\a_{-i}^s} f[k]$
   \STATE $\a_{-i}^s[k{+}1]{=} \text{soft}_{\lambda_s/L_f}(\mathbf{z}^s[k])$
   \STATE $ b_{ii}^s[k+1] =  b_{ii}^s[k]-(1/L_f)\nabla_{b_{ii}^s}f[k] $
   \STATE Update $\a_i^s[k+1]$ via \eqref{eq:zero_pad}
   \ENDFOR
   \STATE $k = k + 1$
   \ENDWHILE
   \STATE $\hat{\A}^s_t = \A^s[k], \hat{\B}^s_t = \B^s[k]$
   \ELSE
       \STATE $ \hat{\A}^s_t=\hat{\A}^s_{t-1}, \hat{\B}^s_t=\hat{\B}^s_{t-1}$
       \STATE  $\alpha_t^s=\alpha_{t-1}^s, \bm\Omega_t^s=\bm\Omega_{t-1}^s,\bar{\Y}_t^s=\bar{\Y}_{t-1}^s$
   \ENDIF
   \ENDFOR
   \ENDFOR
   \RETURN $\{ \hat{\A}^s_t, \hat{\B}^{s}_t \}_{s=1}^S, \{ \hat{\sigma}(t) \}_{t=1}^T$
\end{algorithmic}
\end{algorithm}
%
Algorithm \ref{alg1} summarizes the developed state-dependent topology tracking scheme. Numerical tests indicate that $3-5$ inner ISTA iterations suffice to track the evolving topology remarkably well.

\subsection{Initialization of Algorithm~\ref{alg1}}
\label{subsec:initalg1}
In order to run Algorithm~\ref{alg1}, one needs initial state estimates $\{ \A_0^s, \B_0^s \}_{s=1}^S$. If $\X$ is known to have full row rank, Algorithm~\ref{alg0} can be used to initialize the state matrices as the $S$ cluster centroids. However, this is quite restrictive, and a more general initialization scheme can be obtained with less stringent restrictions on $\X$. For example, the following regularized LS estimator with $\mu > 0$
\begin{equation}
\label{eq:expts}
\underset{  \{ \A, \B : a_{ii}=0, b_{ij} = 0\}}{\text{arg min}} (1/2) 
\| \Y_t - \A \Y_t - \B \X \|_F^2 + \mu \|\A\|_F^2
\end{equation}
yields estimates $ \{ \hat{\A}_t, \hat{\B}_t \}_{t=1}^{T_{\text{init}}}$ per $t$ for a designated initialization interval $t=1,\dots,T_{\text{init}}$ ($T_{\text{init}} \ll T$). 
The initializations $\{ \A_0^s, \B_0^s \}_{s=1}^S$ are then obtained as the $S$ cluster centroids of  $ \{ \hat{\A}_t, \hat{\B}_t \}_{t=1}^{T_{\text{init}}}$. Note that~\eqref{eq:expts} decouples across nodes, and amounts to solving
\begin{equation}
\label{eq:rreg1}
\underset{  \a_{-i}, b_{ii} }{\text{arg min}} \;\; (1/2) 
\|  \y_{i,t} - \Y_{-i,t}^{\top} \a_{-i} - b_{ii} \x_i   \|_2^2 + \mu \| \a_{-i} \|_2^2
\end{equation}
for $i = 1, \dots, N$. Indeed, \eqref{eq:rreg1} admits the following 
per-variable closed-form solutions 
\begin{equation}
\label{eq:rreg2}
\a_{-i} = 
\left( \Y_{-i,t} \Y_{-i,t}^{\top} + 2 \mu \I  \right)^{-1} \Y_{-i,t} \left( \y_{i,t} - b_{ii} \x_i \right)
\end{equation}
and
\begin{equation}
\label{eq:rreg3}
b_{ii} = \frac{ \left( \a_{-i}^{\top} \Y_{-i,t} - \y_{i,t}^{\top}  \right)  \x_i }{\| \x_i  \|_2^2}
\end{equation}
for $i = 1, \dots, N$. Starting with an initial value for $b_{ii}$, one can compute~\eqref{eq:rreg2} and~\eqref{eq:rreg3} in an alternating fashion by fixing one variable and updating the other, until convergence is attained.

\subsection{Tracking slowly-changing state matrices}
\label{subsec:slow}
It has tacitly been assumed that state matrices $\{ \A^s, \B^s  \}_{s=1}^S$ are static, and that temporal dynamics only arise from the switching sequence $\{ \sigma(t) \}_{t=1}^T$. Nevertheless, entries of each of the state matrices may drift slowly over time, motivating algorithms that down-weigh the influence of past data. To this end, (P1) can be modified as follows
\begin{eqnarray}
\label{eq:slow1}
\nonumber \text{(P2)} \;\;\;\;
\underset{ \A^s, \B^s }{\text{arg min}} & &
\frac{1}{2}\sum\limits_{\tau=1}^{t} \beta^{t-\tau} \hat{\chi}_{\tau s} 
\| \Y_{\tau} - \A^s \Y_{\tau} - \B^s\X  \|_F^2 \\ 
\nonumber
& & + \lambda_s \| \A^s \|_1 \\
\text{s. to} 
& &  a_{ii}^s = 0, \; b_{ij}^s = 0, \; \forall  i \ne j.
\end{eqnarray}
where $\beta \in (0,1]$ is a forgetting factor that exponentially down-weighs past data whenever $\beta < 1$.  In order to solve (P2), Algorithm~\ref{alg1} can be readily modified in a reasonably straight-forward manner, and details are omitted here for brevity. 

%
%

\section{Numerical Tests}
\label{sec:experiments}

\subsection{Synthetic Data}
\label{ssec:sdata}
\begin{figure}[!tb]
\begin{minipage}[b]{.49\textwidth}
  \centering
  \includegraphics[width=6.5cm]{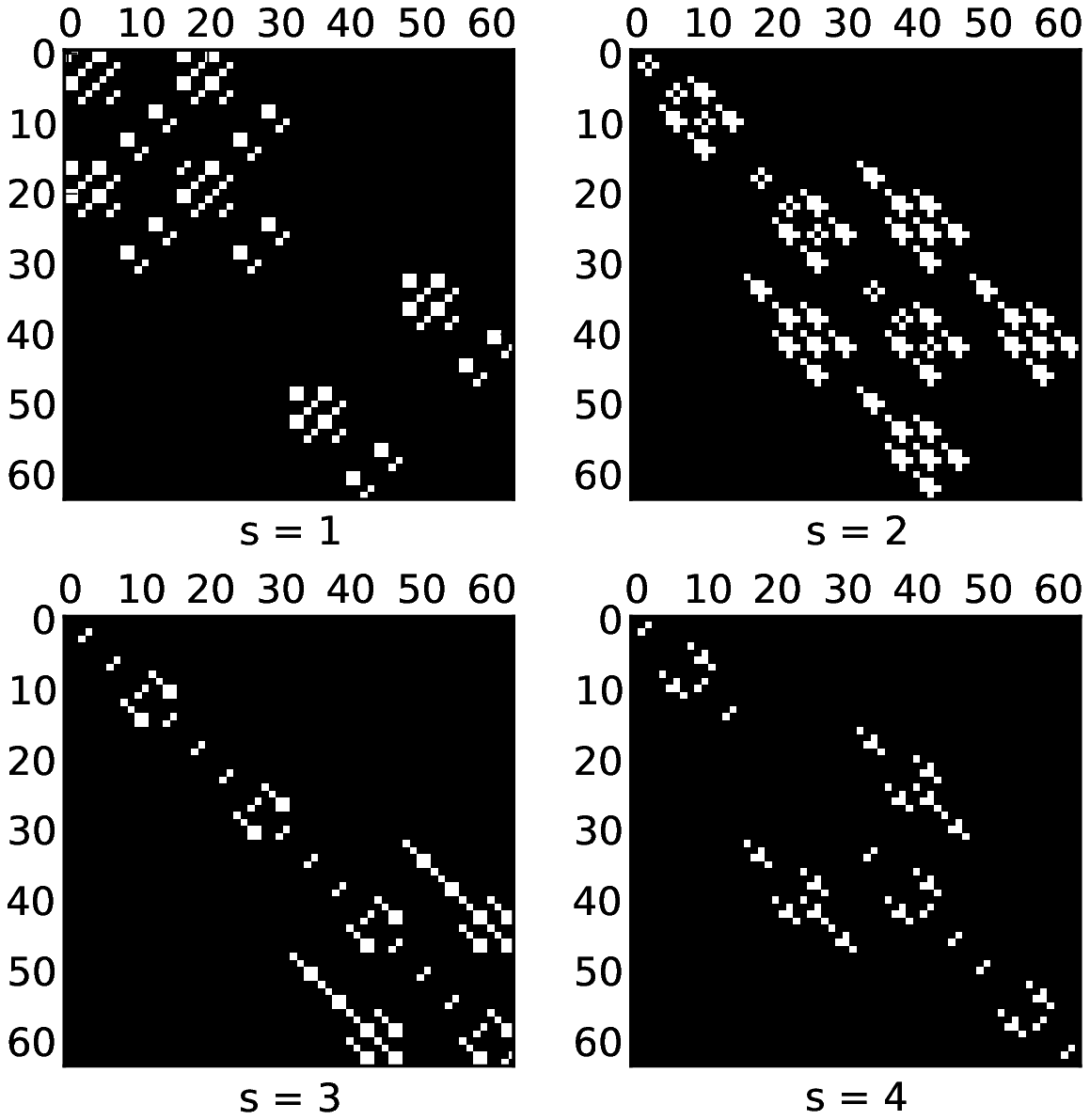}
  \centerline{(a)} 
\end{minipage}
\begin{minipage}[b]{0.49\textwidth}
  \centering
 \includegraphics[width=7.1cm]{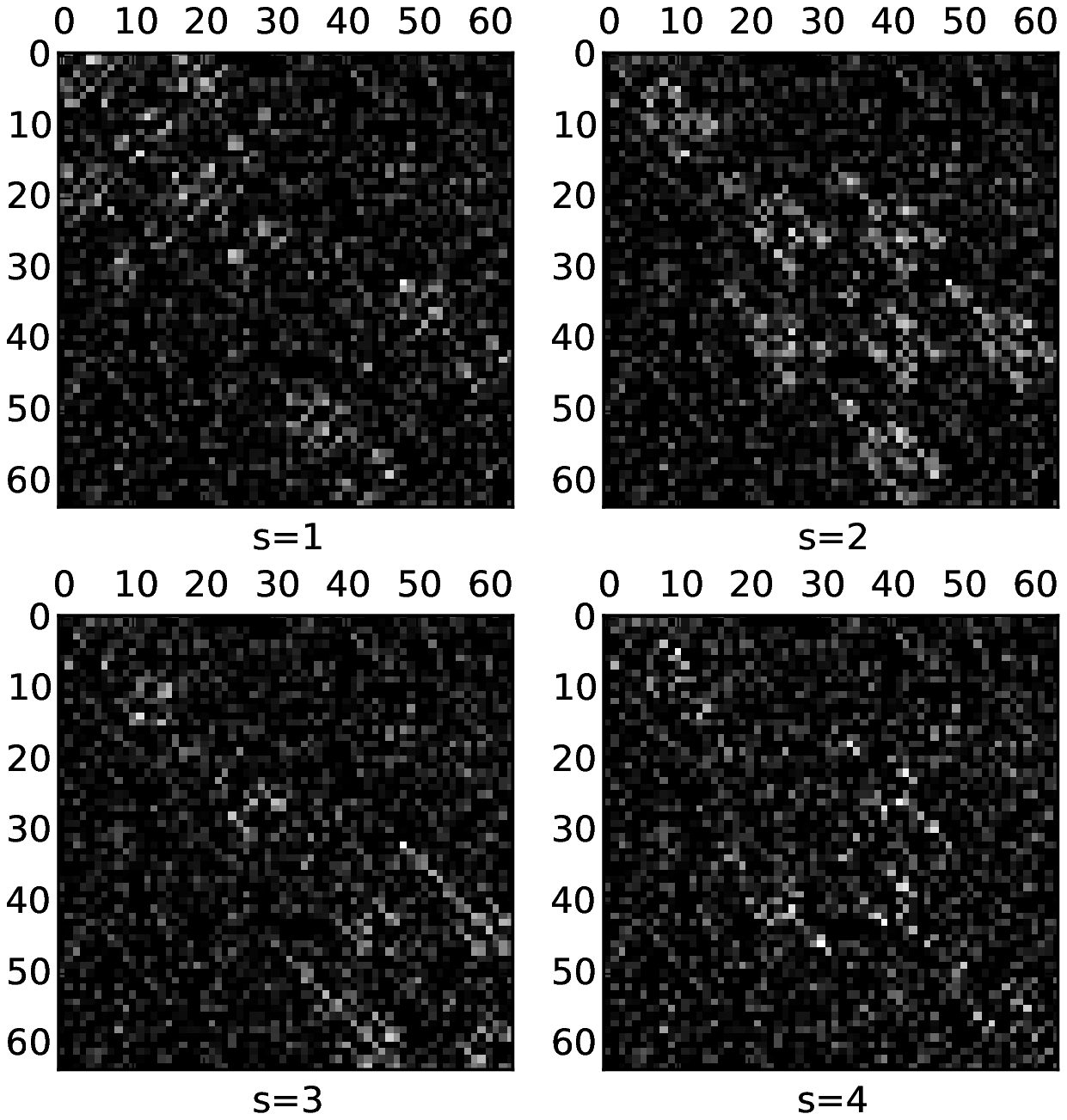}
  \centerline{(b)} 
\end{minipage}
\begin{minipage}[b]{0.49\textwidth}
  \centering
 \includegraphics[width=6.5cm]{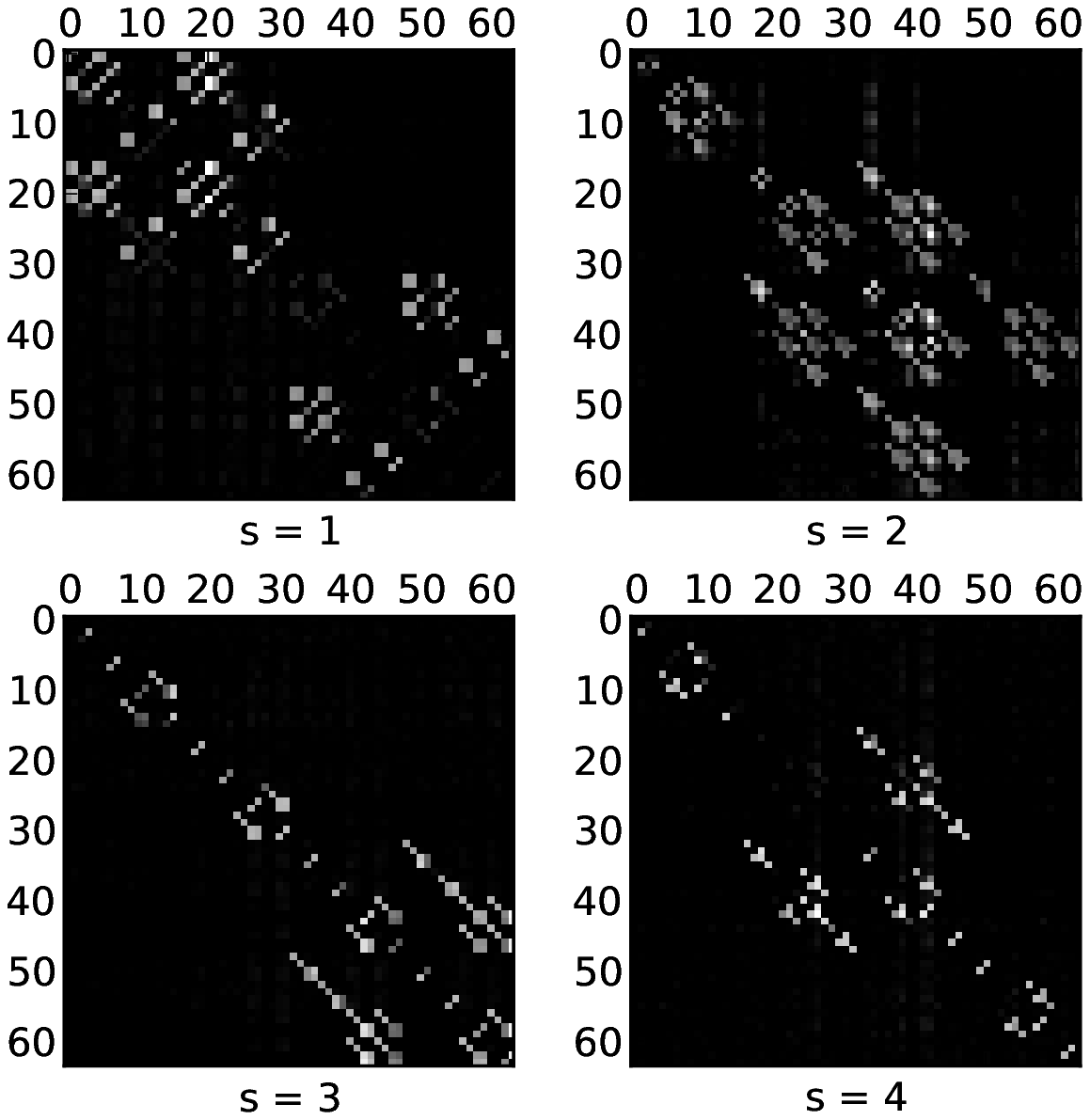}
  \centerline{(c)} 
\end{minipage}
\caption{Adjacency matrices corresponding to: (a) actual switching networks used for the synthetic dataset; (b) state-dependent network topologies inferred by Algorithm~\ref{alg0}; and (c) state-dependent network topologies inferred by Algorithm~\ref{alg1}.}
\label{fig:fig_synth_plots}
\end{figure}

\noindent \textbf{Data generation.} 
To assess the performance of the developed algorithms, the first set of experiments were conducted on synthetic cascade data. Four Kronecker graphs were generated from the following seed matrices~\cite{kron} 
%
\[
\mathbf{H}_1 =  
\begin{pmatrix} 
1 & 1 & 0 & 0 \\ 
1 & 1 & 0 & 0 \\ 
0 & 0 & 0 & 1 \\ 
0 & 0 & 1 & 0 
\end{pmatrix}
\quad
\mathbf{H}_2 = 
\begin{pmatrix}
1 & 0 & 0 & 0 \\
0 & 1 & 1 & 0 \\
0 & 1 & 1 & 1 \\
0 & 0 & 1 & 0
\end{pmatrix}
\]
%
\[
\mathbf{H}_3 = 
\begin{pmatrix}
1 & 0 & 0 & 0 \\
0 & 1 & 0 & 0 \\
0 & 0 & 1 & 1 \\
0 & 0 & 1 & 1
\end{pmatrix}
\quad
\mathbf{H}_4 = 
\begin{pmatrix}
1 & 0 & 0 & 0 \\
0 & 0 & 1 & 0 \\
0 & 1 & 1 & 1 \\
0 & 0 & 0 & 1
\end{pmatrix}.
\]
%
The resulting graph adjacency matrices $\{ \A^s \in \mathbb{R}^{64 \times 64} \}_{s=1}^4$, each encoding a network of $N = 64$ nodes, were obtained by repeated Kronecker products $\A^s = \mathbf{H}_s \otimes \mathbf{H}_s \otimes \mathbf{H}_s$ for $s=1,\dots,4$. Each diagonal entry of $\A^s$ was set to zero, while the number of cascades was set to $C=80$, and time intervals to $T=1,000$. Furthermore, $\X \in \mathbb{R}^{N \times C}$ was constructed with entries sampled from a uniform distribution as $[\X]_{ij} \sim \mathcal{U}[0,3]$. Similarly, the diagonal matrices $\{ \B^s \in \mathbb{R}^{N \times N}  \}_{s=1}^4$ were constructed by sampling entries from  a uniform distribution as $[\B]_{ii} \sim \mathcal{U}[0,1]$. Synthetic cascade data were then generated as 
$\Y_t = (\mathbf{I}_N - \A^{\sigma(t)})^{-1}(\B^{\sigma(t)}\X + \E_t)$, with $\sigma(t)$ sampled uniformly at random from $\mathcal{S} = \{1,2,3,4\}$ per $t$, and  $[\E_t]_{ij} \sim \mathcal{N}(0,0.01)$, for $t=1,\dots,T$. Figure~\ref{fig:fig_synth_plots} (a) depicts the ground-truth adjacency matrices $\{ \A^s \}_{s=1}^4$ used to generate the synthetic cascade data.

\noindent \textbf{Experimental results.} 
First, Algorithm~\ref{alg0} was run with $T_{\text{train}} = 200$, and k-means as the clustering algorithm of choice. Figure~\ref{fig:fig_synth_plots} (b) depicts heatmaps of the resulting adjacency matrices obtained as cluster centroids. Note that Algorithm~\ref{alg0} is devoid of a data-driven thresholding scheme, hence most entries in the recovered adjacency matrix are nonzero. Nevertheless, visual inspection of the heatmaps reveals that larger entries generally correspond to non-zero edge weights as shown in Figure~\ref{fig:fig_synth_plots} (a). Since the synthetic cascade data are noisy, it is not surprising that Algorithm~\ref{alg0}  exhibits suboptimal topology identification. In fact as demonstrated next, Algorithm~\ref{alg1} is markedly superior to Algorithm~\ref{alg0} with respect to coping with noise,  as well as shrinking non-edge entries of $\{ \A^s \}_{s=1}^4$ to zero.

\begin{figure}[!tb]
\centering
\includegraphics[width=0.5\textwidth]{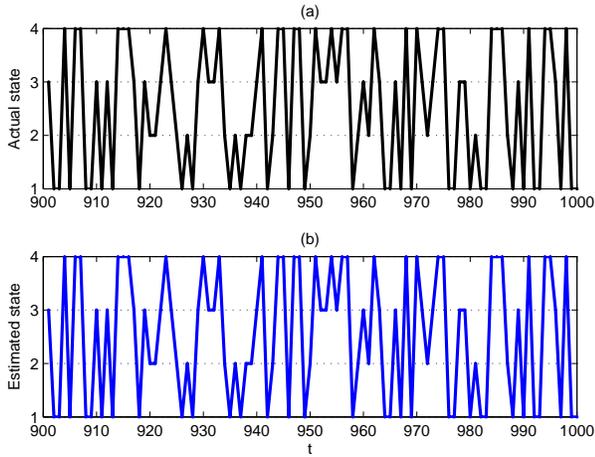}
\caption{Actual (a) and estimated (b) switching sequences plotted from $t=900$ to $t=1,000$.}
\label{fig:fig2}
\end{figure}

Using the initial $50$ cascade samples $\{ \Y_t \}_{t=1}^{50}$, per-interval state-agnostic batch estimates $\{ \hat{\A}_t, \hat{\B}_t \}_{t=1}^{50}$ were obtained by solving~\eqref{eq:expts} with $\mu=0.01$. The estimates $\{ \hat{\A}_t, \hat{\B}_t \}_{t=1}^{50}$ were then clustered via the k-means algorithm, with $k = 4$ clusters, and the corresponding cluster centers were used as the initial values $\{ \hat{\A}^s_0, \hat{\B}^s_0 \}_{s=1}^4$. Algorithm~\ref{alg1} was then run for $t=51,\dots,1,000$, with $\lambda_s = 0.95$, and $s \in \mathcal{S}$. Figure~\ref{fig:fig_synth_plots} (c) depicts heatmaps of the recovered state-dependent adjacency matrices at $T=1,000$. Thanks to sparsity regularization, Algorithm~\ref{alg1} unveils the non-zero support structures of the state matrices with remarkable success. Figure~\ref{fig:fig2} plots the actual and estimated switching sequences from $t=901$ to $t=1,000$, clearly demonstrating the remarkable success in tracking the underlying states.

\begin{figure*}[!tb]
\begin{minipage}[b]{.49\textwidth}
  \centering
  \includegraphics[width=8.5cm]{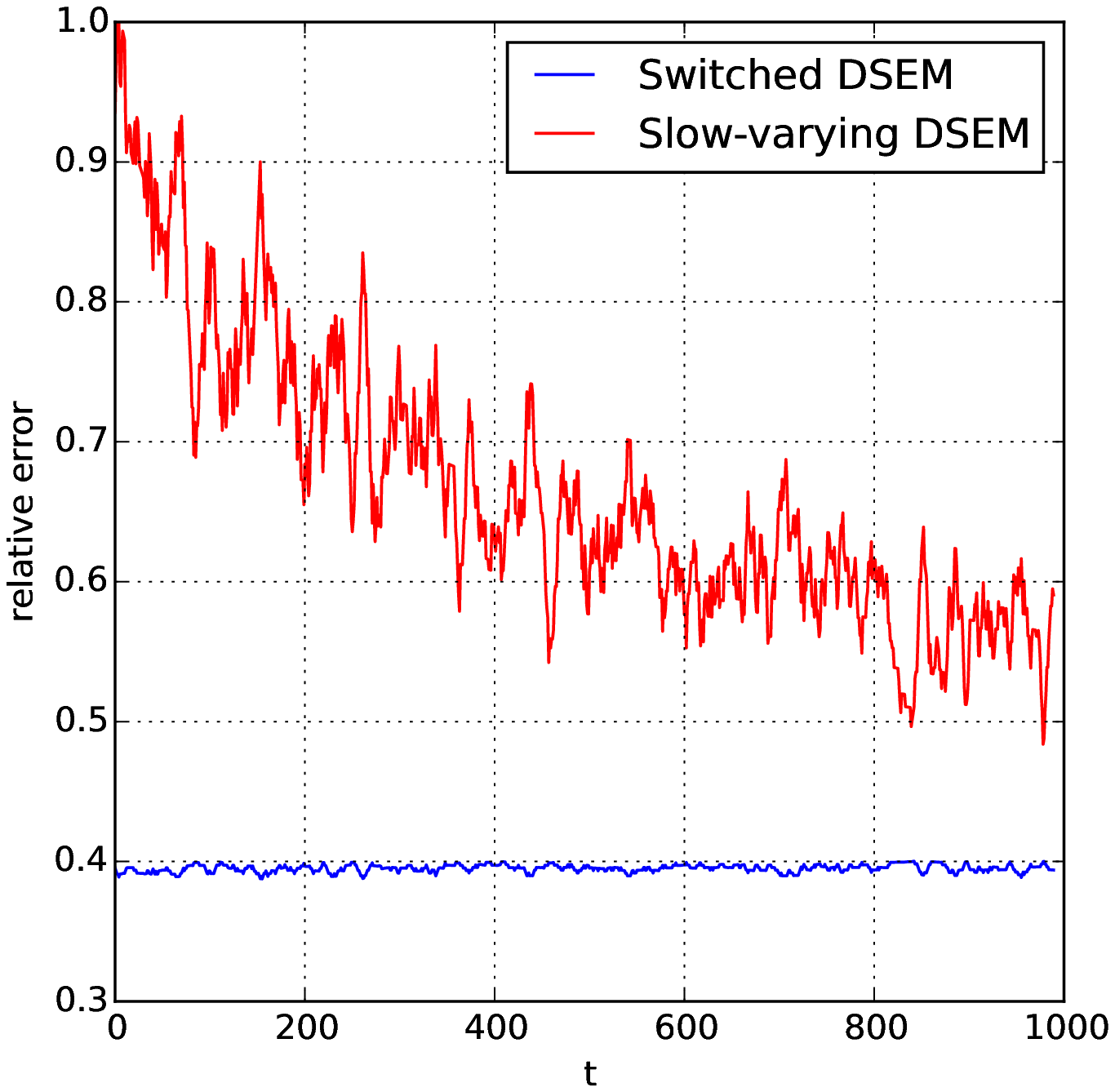}
  \centerline{(a)} 
\end{minipage}
\begin{minipage}[b]{0.49\textwidth}
  \centering
 \includegraphics[width=8.5cm]{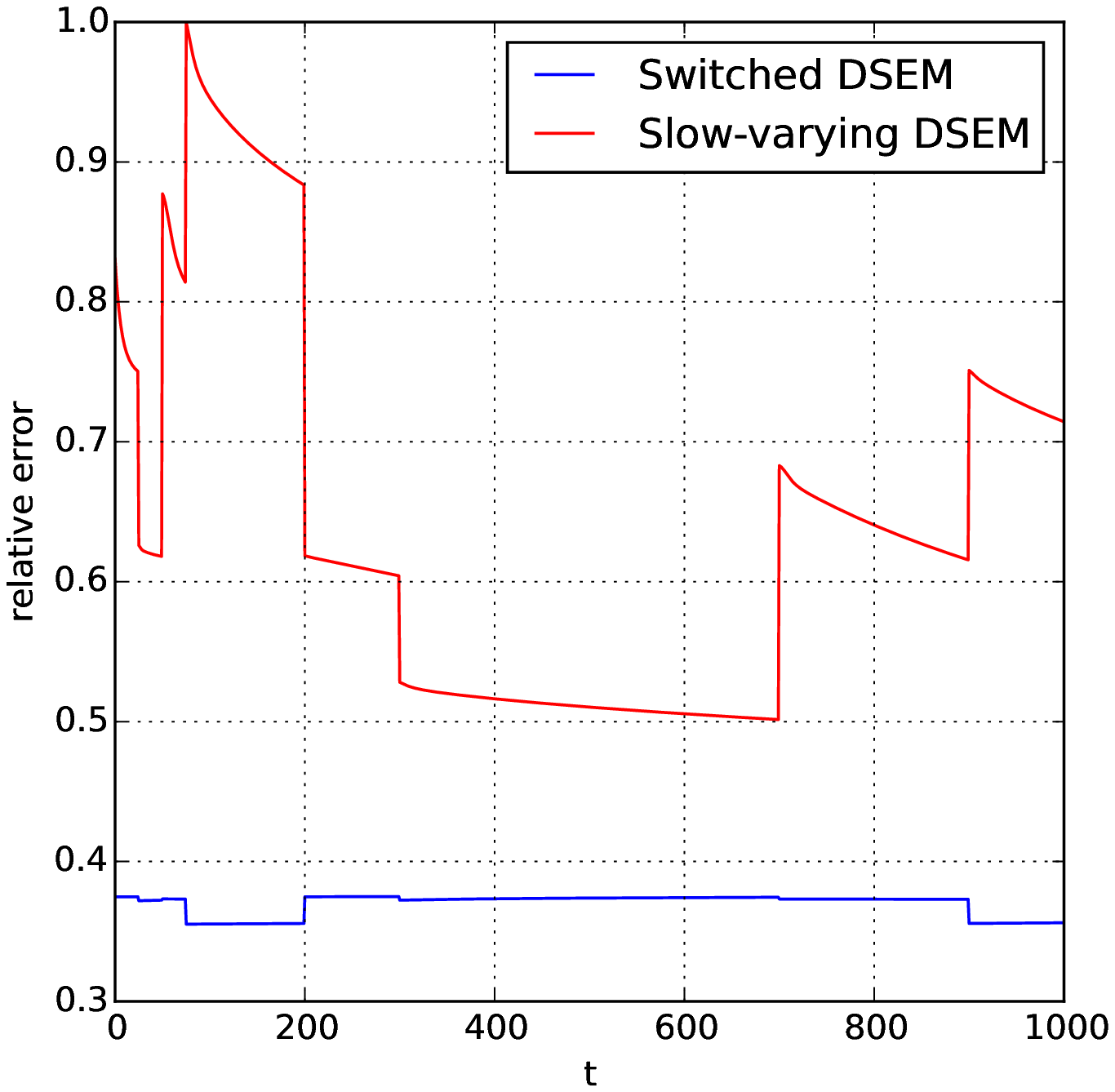}
  \centerline{(b)} 
\end{minipage}
\caption{Comparison of relative errors resulting from tracking the unknown topologies as a switched sequence versus slowly-varying changes: (a) rapidly switching state sequence; and (b) piecewise-constant state sequence.}
\label{fig:fig_synth_comp}
\end{figure*}

Next, the advocated approach was compared with earlier work which models temporal information cascades using dynamic SEMs, with slow topology variations~\cite{baingana_jstsp}. A stochastic gradient descent (SGD) algorithm developed in~\cite{baingana_jstsp} was run with batch initialization on the time-series of cascade data. Per interval $t$, the resulting relative estimation error, $( \| \A^t - \hat{\A}^t \|_F + \| \B^t - \hat{\B}^t \|_F  )/( \| \hat{\A}^t  \|_F  + \| \hat{\B}^t  \|_F )$, was evaluated for both Algorithm~\ref{alg1} and the SGD tracker from~\cite{baingana_jstsp}. Note that in computing the estimation error for Algorithm~\ref{alg1}, $\A^t = \A^{\sigma(t)}, \B^t = \B^{\sigma(t)} $
(similarly for $\hat{\A}^t$ and $\hat{\B}^t$). Figure~\ref{fig:fig_synth_comp} (a) compares the per-interval relative errors of the two approaches. It is clear from the plot that exploiting the prior knowledge that network dynamics arise due to random switching between $S=4$ states yields remarkably superior error performance than the alternative. In fact, the final adjacency matrix $\hat{\A}^T$ does not correspond to a specific state, but is rather an average of the underlying state matrices. However, this is not surprising since the framework advocated in~\cite{baingana_jstsp} exploits slow topology variations, and it is not expected to outperform algorithms developed in the present paper, when topologies potentially jump suddenly between discrete states. 

To this end, a new slowly-varying state sequence was used to control the evolving network topologies, and a new time series of synthetic cascades was generated.  Specifically, the following scheme was used to generate a piecewise-constant sequence $ \{ \sigma(t) \}_{t=1}^{T}$:
\[
\sigma(t) = 
\begin{cases}
1, \quad t \in \{ \{ 1, \dots, 24 \} \cup \{200, \dots, 299 \} \} \\
2, \quad t \in \{ \{ 25, \dots, 49 \} \cup \{300, \dots, 699 \} \} \\
3, \quad t \in \{ \{ 50, \dots, 74 \} \cup \{700, \dots, 899 \} \} \\
4, \quad t \in \{ \{ 75, \dots, 199 \} \cup \{900, \dots, 1,000 \} \}
\end{cases}
\]
while the ground-truth matrices $\{ \A^s, \B^s \}_{s=1}^S$ and $\{ \E_t \}_{t=1}^T$ remained unchanged. Figure~\ref{fig:fig_synth_comp} (a) compares per-interval 
relative errors resulting from running Algorithm~\ref{alg1}, and the SGD algorithm developed in~\cite{baingana_jstsp}. In this case, there are fewer sudden transitions between network states. Nevertheless, leveraging the prior information about the state-based network evolution is beneficial as shown by the plot. Indeed, modeling cascade propagation by the proposed switched dynamic SEM framework leads to better error performance than adopting a state-incognizant approach that exploits the underlying slow variations.
\subsection{Real Cascade Data}
\label{ssec:rdata}
\noindent \textbf{Dataset description.} 
This section presents results of experimental tests conducted on real cascades observed on the web between March $2011$ and February $2012$. 
Blog posts and news articles for memes (popular textual phrases) appearing within on approximately $3.3$ 
million websites were monitored during the observation period. The time  when a given website first mentioned a story related to a specific set of memes was recorded, and the data 
were availed to the public from~[InfoPath]. Cascade infection times were recorded as Unix timestamps in hours (i.e., the number of hours since 
midnight on January $1$, $1970$). Several trending news topics during this period were identified, and cascade data for the top $5,000$ websites 
that mentioned memes associated with them were retained. From this dataset, cascades related to $10$ broad news topics were extracted for 
the present paper's experiments. Table~\ref{tab1} lists the names of these topics, with a correspondingly brief description.
\begin{table*}
\centering
{\renewcommand{\arraystretch}{1.4}
\begin{tabular}{ | c | l | p{13cm} |}
\hline
 & \textbf{Broad news topic} & \textbf{Brief description} \\
\hline \hline
1 & Fukushima & Nuclear accident at Japan's Fukushima nuclear power plant in March 2011 \\
\hline
2 & Kate Middleton & English royal whose wedding took place in April 2011 \\
\hline
3 & Kim Jong-un & Leader of North Korea who rose to prominence upon the death of his father in December 2011  \\
\hline
4 & Osama bin Laden & Infamous terrorist leader who was killed in May 2011 \\
\hline
5 & Amy Winehouse & Famous English singer who died of drug overdose in July 2011 \\
\hline
6 & Rupert Murdoch & Businessman whose media company was involved in a phone hacking scandal revealed in May 2011\\
\hline
7 & Steve Jobs & Technology entrepreneur whose death in October 2011 was followed by many news headlines \\
\hline
8 & Arab spring & Wide-spread politically-charged protests among several Arab nations starting in December 2010 \\
\hline
9 & Strauss Kahn & International Monetary Fund (IMF) director who resigned in May 2011 due to sex assault allegations \\
\hline
10 & Reid Hoffman & Founder of LinkedIn whose stock started trading in May 2011 \\
\hline
 \end{tabular}
 }
\caption{The $10$ broad news topics whose memes constituted the cascades tracked for topology inference.}
\label{tab1}
\end{table*}

For each broad topic in Table~\ref{tab1}, many news stories appeared on blogs and mainstream news websites over the 
observation period. Each story was assigned a list of tuples in the form (website id, timestamp) capturing the time when it 
was first mentioned on a particular website. For the present paper, a popular news story was characterized as a meme
if it propagated to at least $100$ websites, due to repeated mentions. Based on this definition, all news stories that did not
qualify as memes were discarded, and a total of $C = 625$ cascades constituting the selected memes, and their
infection times were retained. Due to this thresholding, the total number of of websites reduced to
$N = 1,131$. The observation period was split into $T = 180$ time intervals $\{ \mathcal{T}_t  \}_{t=1}^{180}$, each corresponding to
approximately $2$ days. Letting $u_{ic}$ denote the Unix infection time of node $i$ by cascade $c$, datum $y_{ic}^t$ was
computed as follows:
\begin{equation}
\label{eq:real_eqn1}
y_{ic}^t =  \log_{10} \left( u_{ic} - \underset{u_{ic} \in \mathcal{T}_t}{\text{min}} \; u_{ic}  \right)
\end{equation}
only if $u_{ic} \in \mathcal{T}_t$, otherwise $y_{ic}^t $ is set to a very large value, namely
$y_{ic}^t  = 2 + \log_{10} ( \underset{i,c}{\text{max}} \; u_{ic} ) $, as a surrogate for infinity. 

Entries of $\X$ capture prior knowledge about the susceptibility of each node to each
contagion. For instance, the entry $x_{ic}$ could denote the online search rank of website $i$ 
for a search keyword associated with contagion $c$. In the absence of such website ranking
information, the present paper resorted to an alternative approach involving
assignment of susceptibilities from the entire corpus of the cascade data. 
First, five broad categories of memes were identified 
as \emph{politics}, \emph{entertainment}, \emph{sports}, \emph{business}, and \emph{technology},
and indexed by $k = 1, \dots, 5$. Each group of news topics was manually labeled using these 
five prescribed categories. Next, for each website $i$, $\gamma_{ik}$ was computed as follows
\[
\gamma_{ik} := \frac{\text{number of category $k$ cascades that infected node $i$}}
{\text{total number of cascades that infected node $i$}}
\]
for all $k$ categories. Entry $x_{ic}$ was then set to $\gamma_{ik}$ if
cascade $c$ belonged to category $k$.

\noindent \textbf{Experimental results.} Since $S$ is unknown for the real dataset, the first part of this 
experimental test entailed approximating its value from the data. Setting $\mu = 0.15$, the initialization scheme in 
\eqref{eq:expts} was run in batch mode over $60$ intervals. The resulting estimates of
$\{ \hat{\A}_t, \hat{\B}_t \}_{t=1}^{60}$ were then clustered using $S$-means, with
$S=1,\dots,10$. For each $S$, the total intra-cluster distance was evaluated upon convergence as
\begin{equation}
\label{eq:int_clust}
\delta(S) := \log_{10} \left\lbrace \sum\limits_{s=1}^S  \sum\limits_{t=1}^{60}  \hat{\chi}_{ts} \| \hat{\boldsymbol{\theta}}^t 
- \hat{\boldsymbol{\theta}}^s  \|_2^2   \right\rbrace
\end{equation}
on a logarithmic scale, with the estimates $\hat{\chi}_{ts},  \hat{\boldsymbol{\theta}}^t$, and  $\hat{\boldsymbol{\theta}}^s $ defined
earlier. Figure~\ref{fig:fig_cluster_disp} plots $\delta(S)$ for $S = 1, \dots, 10$, depicting a 
substantial decrease from $S=2$ to $S=3$. Subsequent values of $S$ do not markedly reduce $\delta(S)$. 
From this plot, $S = 3$ was selected as the underlying number of network states.

Next, Algorithm~\ref{alg1} was run with the pre-processed cascade data as inputs. Note that Algorithm~\ref{alg1}
requires the sparsity-promoting regularization parameters $\{  \lambda_s \}_{s=1}^S$ as inputs. In this experiment,
a uniform value was adopted for all states, i.e., $\lambda_1 = \lambda_2  = \lambda_3 = \lambda$. In the absence
of ground-truth state topologies, selection of $\lambda$ is rather challenging, and it remains an open question 
for future work. Nevertheless, a heuristic approach inspired by typical properties of real-world large-scale networks
was adopted for the present paper. Over the last $10-15$ years, several studies in network science have unveiled
remarkable universal properties that seem to underlie most real networks. For example, networks generally exhibit
the so-termed ``small-world'' property, and their degree distributions often follow power laws~\cite{watts_book}. 

Acknowledging these as universal laws inherent to most networks, it is possible to constrain the search space of the most likely network 
topologies uncovered by Algorithm~\ref{alg1}. In this experiment, $\lambda$ was selected in such a way that the 
average shortest path length between any pair of nodes is $\mathcal{O}(\log N)$, which is consistent with the
``small-world'' phenomenon. With $N = 1,131$, the goal was to select $\lambda$ so that the resulting 
average shortest path length was within the interval $[6,10]$. For each $\lambda$, Algorithm~\ref{alg1} was run
and the average shortest path lengths were computed per network state, and then averaged over all states. 
Figure~\ref{fig:fig_lambda_selection} plots the average path length as $\lambda$ is varied between $0$ and $100$. The plot
also depicts the corresponding average network diameters, which can also be used as a validation metric. It turns
out that setting $\lambda = 10$ for $s \in \{1,2,3\}$ led to an average shortest path length of approximately $7$, which
is indicative of an underlying ``small-world'' property. Table~\ref{tab2} summarizes the per-state average clustering 
coefficient, network diameter, average number of neighbors, and the average shortest path length when $\lambda=10$.
\begin{figure*}[!tb]
\begin{minipage}[b]{.33\textwidth}
  \centering
  \includegraphics[width=6.5cm]{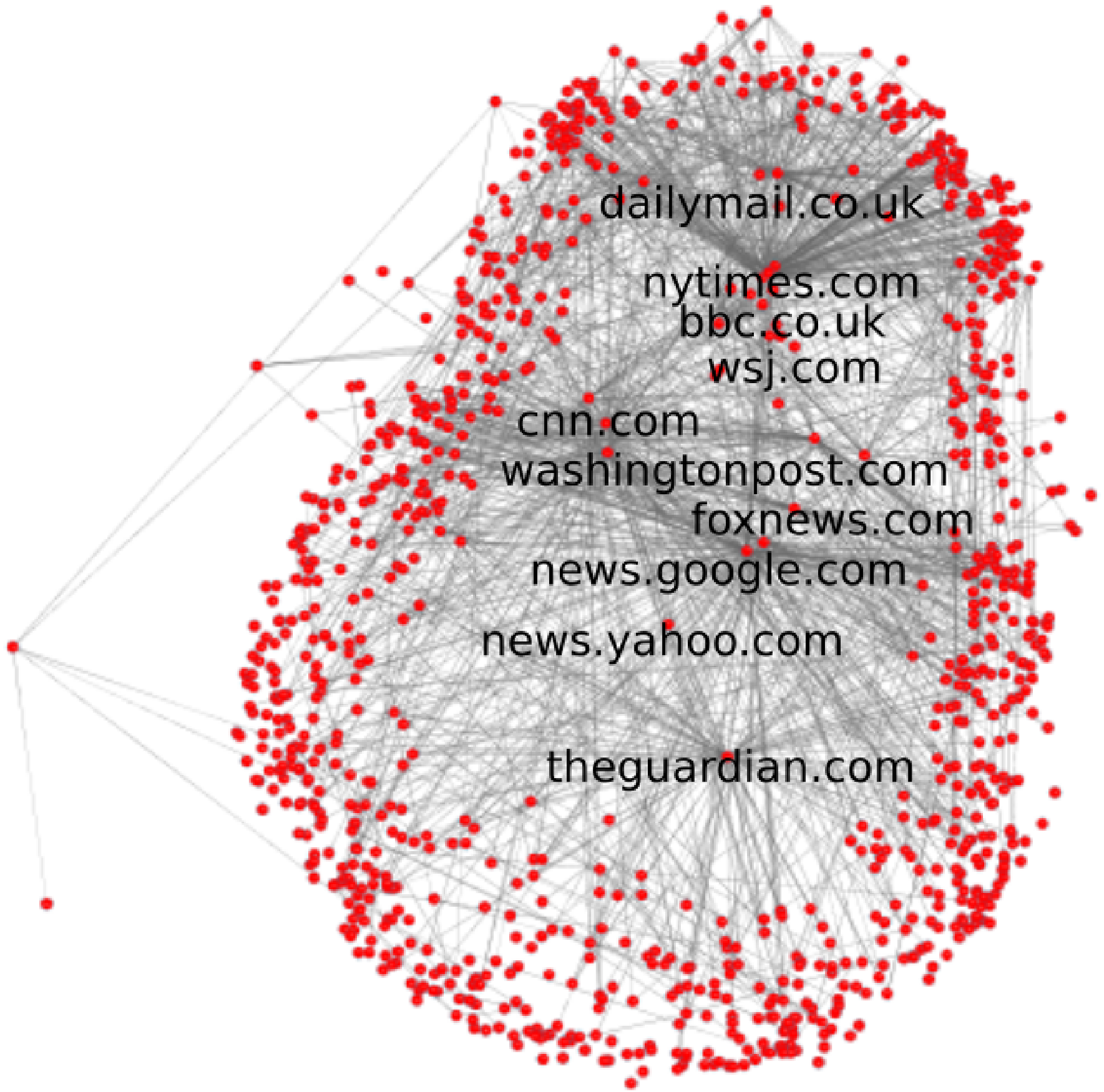}
  \centerline{(a) $s=1$} 
\end{minipage}
\begin{minipage}[b]{0.33\textwidth}
  \centering
 \includegraphics[width=6.5cm]{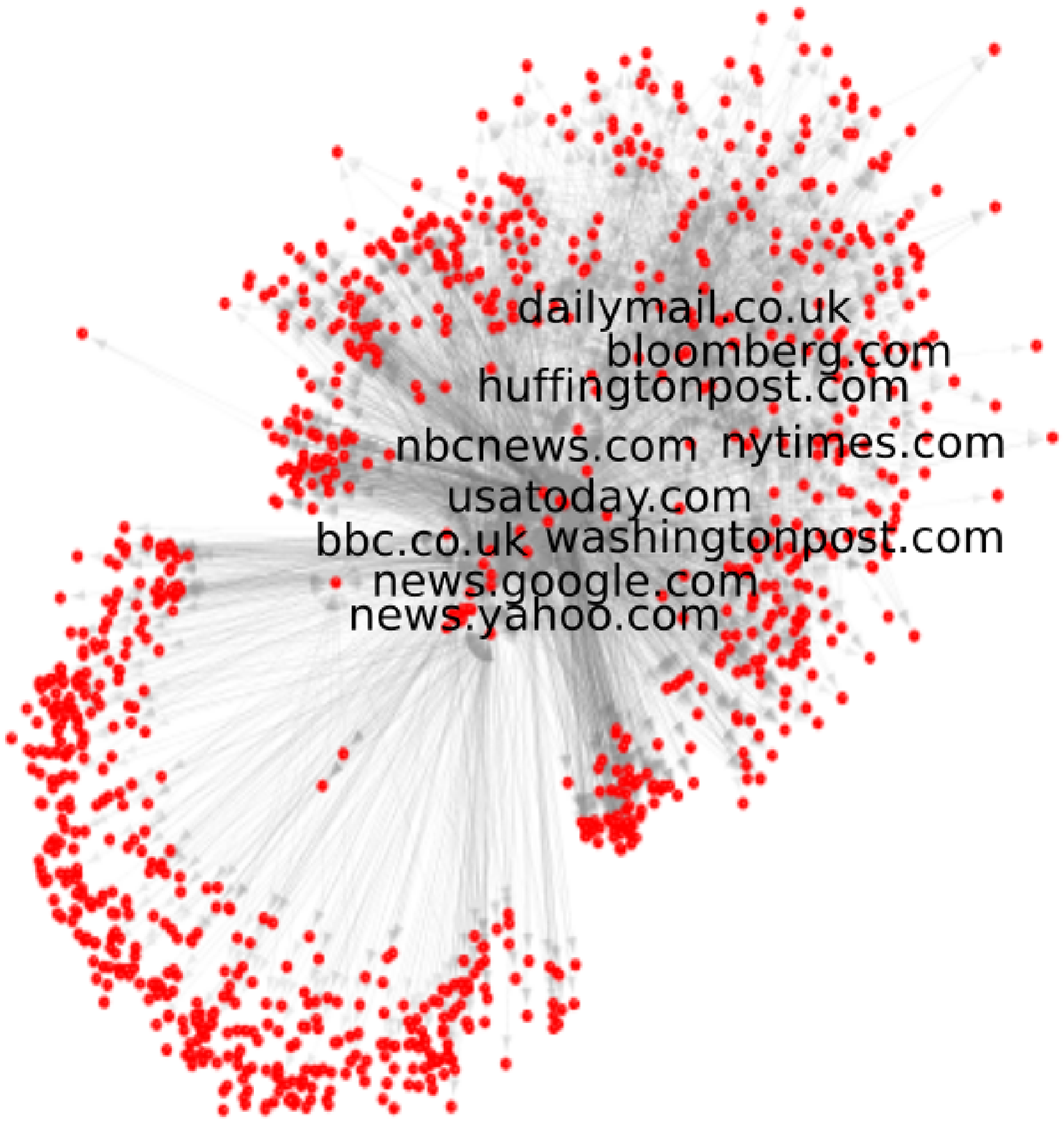}
  \centerline{(b) $s = 2$} 
\end{minipage}
\begin{minipage}[b]{0.33\textwidth}
  \centering
 \includegraphics[width=6.5cm]{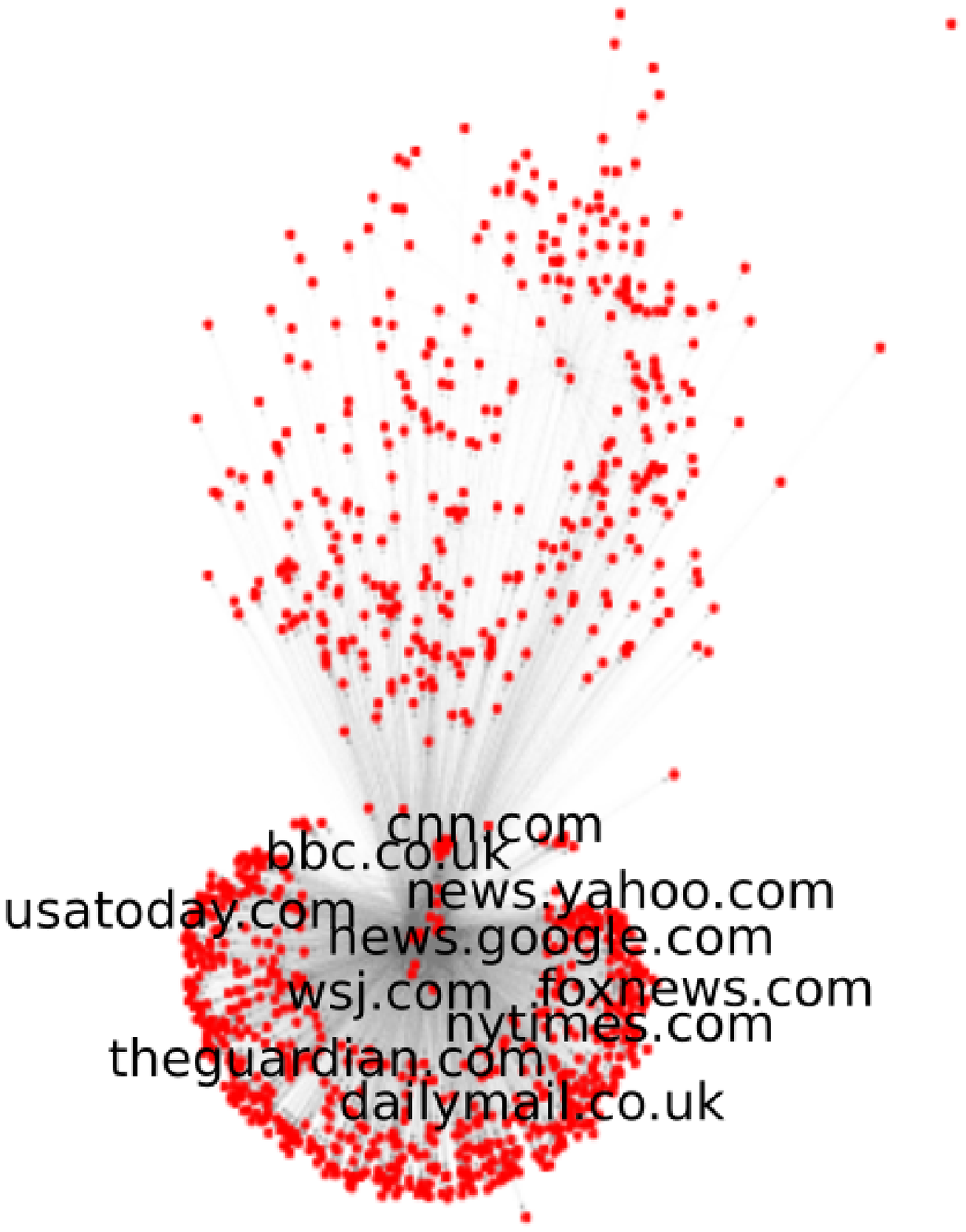}
  \centerline{(c) $s =3$} 
\end{minipage}
\caption{State-dependent network topologies inferred from real cascades annotated by the top $10$ ranked websites, in order of decreasing 
out-degree. Interestingly, Algorithm~\ref{alg1} reveals that the most influential websites turn out to be 
well-known media outlets and popular web-based news aggregators.}
\label{fig:fig_real_nets}
\end{figure*}
\begin{figure}[!tb]
\centering
\includegraphics[width=0.5\textwidth]{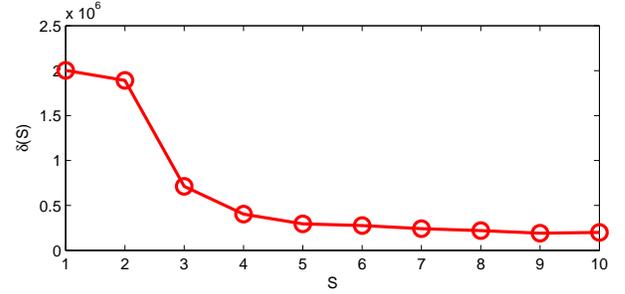}
\caption{Total intra-cluster distances plotted for different values of $S$, with $S=3$ selected as the 
optimal size of the state space in the real web cascades dataset.}
\label{fig:fig_cluster_disp}
\end{figure}
\begin{figure}[!tb]
\centering
\includegraphics[width=0.5\textwidth]{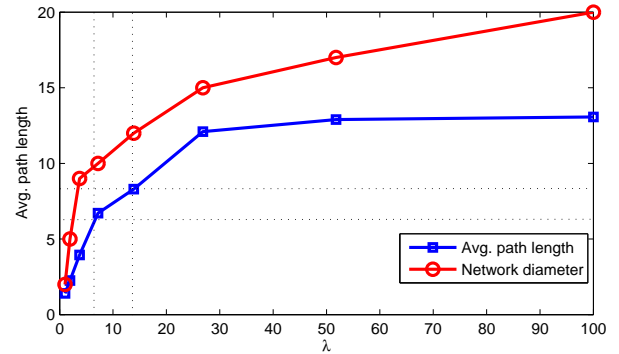}
\caption{Empirical selection of $\lambda$ guided by the resulting average shortest path lengths in the inferred networks. With
$\lambda = 10$, the inferred network topologies exhibit the ``small-world'' property with average path lengths between $6-10$ hops.}
\label{fig:fig_lambda_selection}
\end{figure}
\begin{figure}[!tb]
\centering
\includegraphics[width=0.5\textwidth]{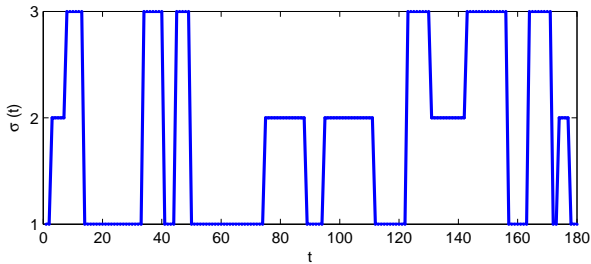}
\caption{Inferred switching sequence from the real-world web cascade data. It turns out that network 
dynamics are governed by piecewise-constant state variations.}
\label{fig:fig_real_state_seq}
\end{figure}
\begin{table}
\centering
{\renewcommand{\arraystretch}{1.4}
\begin{tabular}{ | l | c | c | c |}
\hline
 \textbf{Network state} & \textbf{1} & \textbf{2} & \textbf{3} \\
\hline \hline
\textbf{Average clustering coefficient} & $0.367$ & $0.216$  &  $0.218$ \\
\hline
\textbf{Network diameter} & $18$ & $13$ & $8$  \\
\hline
\textbf{Average number of neighbors} & $134.164$ & $81.132$ & $25.738$  \\
\hline
\textbf{Average shortest path length} & $9.34$ & $7.66$ & $4.51$ \\
\hline
\end{tabular}
}
\caption{Simple summary statistics extracted from the inferred network state topologies with $\lambda=10$.}
\label{tab2}
\end{table}

Figure~\ref{fig:fig_real_nets} depicts visualizations of the state-dependent network topologies inferred from 
the cascade data using Algorithm~\ref{alg1}. Each drawing is annotated with the top $10$ websites when ranked  
in order of decreasing out-degree. It is interesting to note that this list is dominated by websites corresponding to
the most influential news outlets in the English-speaking world e.g., \emph{nytimes.com}, \emph{cnn.com}, \emph{bbc.co.uk}, 
and \emph{theguardian.com}. Furthermore, it turns out that popular news aggregators, such as \emph{news.yahoo.com} 
and  \emph{news.google.com}, are competitive with traditional media powerhouses in terms of influencing
the propagation of news over the web. Although this analysis only considers English-based websites, these experimental 
results on real-world cascades are strongly corroborated by prior expectations about the most influential enablers of
information propagation over the world-wide web. Finally, Figure~\ref{fig:fig_real_state_seq} depicts the switching 
sequence for the network topologies over the entire observation period. It turns out that evolution of the network topology 
adheres to piecewise constant state variations. 

%
%

\section{Conclusion}
\label{sec:conclusion}
A switched dynamic SEM was proposed for tracking the evolution of dynamic networks from information cascades when topologies  arbitrarily switch between discrete states. It was shown that presence of exogenous influences is critical to guarantee model identifiability, and one can even identify the underlying network topologies with fewer cascade measurements by leveraging edge sparsity. Recognizing that identification of the unknown network states and the switching sequence is computationally challenging, the present paper assumed that the number of states are known a priori, and advocated an exponentially-weighted LS estimator capitalizing on edge sparsity. 
A recursive two-step topology tracking algorithm leveraging advances in proximal gradient optimization was developed. Per interval, the first step estimates the active state by minimizing the a priori prediction error, while the second step recursively updates only the matrices corresponding to the estimated state. 

Experiments on synthetically-generated cascades demonstrated the effectiveness of the advocated approach in jointly tracking the switching sequence, while identifying the state-dependent network topologies. Numerical tests were also conducted on real cascades of popular news over the web, collected over one year. The goal was to identify the underlying topological structure of causal influences between news websites and blogs. Interestingly, websites exhibiting the highest influence (as measured by the out-degree) turned out to be recognizable news outlets and popular news aggregators. Future directions include investigation of more efficient approaches for tracking the switching sequence, identification of the number of states, and selection of sparsity-promoting parameters, especially when ground-truth network topologies are unknown.

%
%

\begin{appendices}

\section{Proof of Proposition~\ref{prop:ml2}}
\label{app:proof1}
In order to construct the proof for Proposition~\ref{prop:ml2}, it is necessary to state and prove Lemmas~\ref{lemma1} and~\ref{lemma2}.

\begin{lemma}
\label{lemma1}
Under (as5), matrix $\left( \I - (\A^{\sigma(t)})^{\top} \right)$ is invertible with probability one.
\end{lemma}

\noindent \textbf{Proof:} Note that $\text{det} \left( \I - (\A^{\sigma(t)})^{\top} \right)$
is a polynomial whose variables are the nonzero entries of $\A^{\sigma(t)}$. Consequently, for
some constants $\{ \alpha_{ij} \}_{i,j}$, the probability that $\left( \I - (\A^{\sigma(t)})^{\top} \right)$ is
not invertible can be equivalently written as 
\begin{eqnarray}
\label{eq:app1}
\nonumber
\text{Pr} \{ \text{det} ( \I - (\A^{\sigma(t)})^{\top} ) = 0  \} 
=&{ \text{Pr}} \{ \prod\limits_{  i, j  } ( a_{ij}^{\sigma(t)} - \alpha_{ij} ) = 0  \} \\
\nonumber
= &{ \text{Pr}}\{ \bigcup_{i, j} ( a_{ij}^{\sigma(t)} = \alpha_{ij} )  \} \\
 \leq & \sum\limits_{i,j} \text{Pr} ( a_{ij}^{\sigma(t)} = \alpha_{ij} ).
\end{eqnarray}
Since $a_{ij}^{\sigma(t)}$ is drawn from a continuous distribution (as5), it follows that $\text{Pr} ( a_{ij}^{\sigma(t)}{=}\alpha_{ij}  ) {=} 0$,
implying that
$\text{Pr} \{ \text{det} ( \I - (\A^{\sigma(t)})^{\top} ) {\neq} 0  \} {=} 1$. $\text{      }  \blacksquare$

\begin{lemma}
\label{lemma2}
Under (as4)-(as6), assuming $\text{kr}\left( \X^{\top} \right) \geq 2K + 1$, implies that 
\[ 
\text{kr} \left( \Y_t^{\top} \right) = \text{kr} \left( \X^{\top} \B^{\sigma(t)} \left( \I 
- \left( \A^{\sigma(t)} \right)^{\top}  \right)^{-1}  \right) \geq 2K+1
\] 
with probability one.
\end{lemma}

\noindent \textbf{Proof:} Defining $\bar{\X}^{\top} := \X^{\top} \B^{\sigma(t)}$, note 
that $\text{kr} \left( \bar{\X}^{\top} \right) = \text{kr} \left( \X^{\top} \B^{\sigma(t)} \right)$. 
Let $ \bar{\X}^{\top}_{\mathcal{Q}} $ denote a submatrix of $ \bar{\X}^{\top}$,
formed by selecting  an arbitrary subset of columns indexed by the set 
$ \mathcal{Q} \subset \{ 1, \dots, N \}$  whose cardinality is $2K+1$. Given 
that $\text{kr} \left( \bar{\X}^{\top} \right) \geq 2K+1$, there exists a set of rows  of $\bar{\X}^{\top}_{\mathcal{Q}}$
indexed by $ \mathcal{P} \subset \{ 1, \dots, C \}$ with cardinality $2K+1$,
such that the resulting square submatrix $\bar{\X}^{\top}_{\mathcal{Q}, \mathcal{P}}$
has full rank.

Since from Lemma~\ref{lemma1}, we have that $( \I { - } (\A^{\sigma(t)})^{\top} )$ is invertible, it holds that
$ \Y_t^{\top} { = } \bar{\X}^{\top} \left( \I { - }  ( \A^{\sigma(t)} \right)^{\top} )^{-1}   $,
or $ \Y_t^{\top} { = } \bar{\X}^{\top} \Phib $, where $\Phib { := } ( \I 
- ( \A^{\sigma(t)} )^{\top} )^{-1}$. Collecting the columns and rows of $\Y_t^{\top}$
indexed by $\mathcal{Q}$ and $\mathcal{P}$ into $(\Y^t_{\mathcal{Q}, \mathcal{P}})^{\top}$, it follows that
$(\Y^t_{\mathcal{Q}, \mathcal{P}})^{\top} = \bar{\X}^{\top}_{\mathcal{P}} \Phib_{\mathcal{Q}}$,
where $ \bar{\X}^{\top}_{\mathcal{P}}$ collects rows of $\bar{\X}^{\top}$ indexed by $\mathcal{P}$,
and $\Phib_{\mathcal{Q}}$ collects columns of $\Phib$ indexed by $\mathcal{Q}$. Equivalently,
one can write  $(\Y^t_{\mathcal{Q}, \mathcal{P}})^{\top} = \bar{\X}^{\top}_{\mathcal{Q}, 
\mathcal{P}} \Phib_{\mathcal{Q}, \mathcal{Q}}$, where only the corresponding columns 
of $\bar{\X}^{\top}_{\mathcal{P}}$ and rows of $\Phib_{\mathcal{Q}}$ are retained.

By Cramer's rule for matrix inversion, entry $\phi_{ij}$ of $\Phib$ can be written as
$\psi_{ij} (\A^{\sigma(t)})/ \text{det}(\I - \A^{\sigma(t)})$, where  $\psi_{ij} (\A^{\sigma(t)})$
denotes a polynomial function of degree at most $(N-1)$, with nonzero entries of
$\A^{\sigma(t)}$ as variables. Furthermore, the entries of $(\Y^t_{\mathcal{Q}, \mathcal{P}})^{\top}$
are linear combinations of $N$ entries of $\Phib_{\mathcal{Q}}$ , while the determinant of
$ (\Y^t_{\mathcal{Q}, \mathcal{P}})^{\top} $ is a polynomial $ \psi_{\text{det}}
 ((\Y^t_{\mathcal{Q}, \mathcal{P}})^{\top}) $ of degree $2K+1$, with its entries as
 variables. As a result, one can write the determinant of $(\Y^t_{\mathcal{Q}, \mathcal{P}})^{\top}$
 as 
 \begin{equation}
 \label{eq:app2}
 \text{det}\left( (\Y^t_{\mathcal{Q}, \mathcal{P}})^{\top}  \right) = 
 \frac{\bar{\psi}_{\text{det}} \left( \A^{\sigma(t)} \right) }{\text{det} \left( \I - \A^{\sigma(t)}  \right)^{(2K+1)N}}
 \end{equation}
where $\bar{\psi}_{\text{det}} \left( \A^{\sigma(t)} \right) $ is a polynomial of degree at most
$(2K+1)N(N-1)$ with nonzero entries of $ \A^{\sigma(t)}$ as variables, and is formed by
the composition of the polynomials $\psi_{ij} (\A^{\sigma(t)})$ and linear combinations of entries of
$\Phib_{\mathcal{Q}}$.

It can be shown that $\bar{\psi}_{\text{det}} \left( \A^{\sigma(t)} \right)$ is not identically zero as 
follows. Since $\A^{\sigma(t)} = \mathbf{0}$ implies that $\Phib = \I$, then 
$\text{det}\left( (\Y^t_{\mathcal{Q}, \mathcal{P}})^{\top}  \right) = 
\text{det} (\bar{\X}^{\top}_{\mathcal{Q}, \mathcal{P}}) \neq 0$, because 
$\bar{\X}^{\top}_{\mathcal{Q}, \mathcal{P}}$ is full rank by construction. Consequently,
$\bar{\psi}_{\text{det}} \left( \A^{\sigma(t)} \right)$ depends on nonzero entries of
$\A^{\sigma(t)}$, which are drawn from a continuous distribution. Similar to the 
argument made in the proof of Lemma~\ref{lemma1}, the probability
that $\bar{\psi}_{\text{det}} \left( \A^{\sigma(t)} \right) = 0$ is zero. Since 
$(\Y^t_{\mathcal{Q}, \mathcal{P}})^{\top} $ has full rank, then 
$\text{kr} \left( \Y_t^{\top} \right) {  \geq  } 2K+1$ with probability one. $\blacksquare$

\noindent \textbf{Proposition 2:}
\textit{Suppose data matrices $\Y_t$ and $\X$ adhere to \eqref{eq:id0} with $a_{ii}^{\sigma(t)}=0 $,
$b_{ii}^{\sigma(t)} \neq 0, \;\; \forall \; i, t$, and   $b_{ij}^{\sigma(t)} = 0 \;\; \forall \; i \neq j, t$. 
If (as1)-(as6) hold, and $p_s > 0 \;\; \forall s$, then  $\left\lbrace \A^s, \B^s \right\rbrace_{s=1}^S$ 
can be uniquely identified with probability one as $T \rightarrow \infty$}.

\noindent \textbf{Proof:} First rewrite~\eqref{eq:id0} as 
\begin{equation}
\label{eq:app3}
\Y_t^{\top} \mathbf{F}_t = \X^{\top} 
\end{equation}
where $\mathbf{F}_t := (\I - (\A^{\sigma(t)})^{\top}) ( \B^{\sigma(t)} )^{-1} $. Note that 
both $\A^{\sigma(t)}$ and $\B^{\sigma(t)}$ can be uniquely determined from $\mathbf{F}_t$.
Specifically, with $a_{ii}^{\sigma(t)} = 0 \;\; \forall i$, then $\text{Diag} (\mathbf{F}_t) = ( \B^{\sigma(t)} )^{-1}$,
implying that $\B^{\sigma(t)} = (\text{Diag} (\mathbf{F}_t))^{-1}$ and $\A^{\sigma(t)} = \I - \B^{\sigma(t)} \mathbf{F}_t^{\top}$.
As a result, identifiability of $\mathbf{F}_t$ implies identifiability of both $\A^{\sigma(t)}$ and $\B^{\sigma(t)}$, and the
argument is by contradiction.

If there exists $\bar{\mathbf{F}}_t$ satisfying~\eqref{eq:app3}, then $\Y_t^{\top} \tilde{\mathbf{F}}_t = \mathbf{0}$,
where $\tilde{\mathbf{F}}_t := \mathbf{F}_t - \bar{\mathbf{F}}_t$. Since the columns of $\tilde{\mathbf{F}}_t$
have at most $2K+1$ nonzero entries each, and $\text{kr} \left( \Y_t^{\top} \right) \geq 2K+1$ (Lemma~\ref{lemma2}), it 
follows that $\tilde{\mathbf{F}}_t = \mathbf{0}$. Hence, $\mathbf{F}_t = \bar{\mathbf{F}}_t$ which is a contradiction. 
Consequently, $\mathbf{F}_t$ is uniquely identifiable, and so are  $\A^{\sigma(t)}$ and $\B^{\sigma(t)}$ for all $t$.

Recalling that $\A^{\sigma(t)} =  \sum\limits_{s=1}^S \chi_{ts} \A^s $ and $\B^{\sigma(t)} =  \sum\limits_{s=1}^S  \chi_{ts} \B^s $, 
the unique solution from~\eqref{eq:app3} 
coincides with a specific pair in the set $\left \lbrace \left(\A^s, \B^s \right) \right \rbrace_{s=1}^S$ per interval $t$. 
Following a similar argument to the proof of Proposition~\ref{prop:ml},
recall that $\chi_{ts}$ are Bernoulli random variables with $p_s = \text{Pr} ( \chi_{ts} = 1  )$ per $t$. Hence,
the number of times state $s$ is activated over $T$ time intervals follows a binomial distribution. 
With $T_s := \sum\limits_{t=1}^T \chi_{ts}$  denoting the total number of activations of state $s$ over $t = 1, \dots, T$, it turns out 
that $\text{Pr}(T_s \geq 1)  =  1 - \text{Pr}(T_s = 0) =  1 - (1 - p_s)^T$.
Given that $\underset{T \rightarrow \infty}{\text{lim}} \; \text{Pr}(T_s \geq 1) = 1$ only if $p_s > 0$, 
all states can be uniquely identified with probability $1$ as $T \rightarrow \infty$, whenever 
$p_s > 0 \;\; \forall s$. $\text{ } \blacksquare$

\end{appendices}

\bibliographystyle{IEEEtranS}
\bibliography{IEEEabrv,net}

\end{document}